\documentclass[sigconf]{aamas}

\usepackage{balance} 

\usepackage{hyperref}
\usepackage{url}
\usepackage[ruled,vlined]{algorithm2e}
\usepackage{algorithmic}
\usepackage{graphicx}
\usepackage{float}
\usepackage{caption}
\usepackage{subcaption}
\usepackage{booktabs}
\usepackage{wrapfig}
\usepackage{multicol}
\usepackage{amsmath}
\usepackage{bm}

\setcopyright{ifaamas}
\acmConference[AAMAS '23]{Proc.\@ of the 22nd International Conference
on Autonomous Agents and Multiagent Systems (AAMAS 2023)}{May 29 -- June 2, 2023}
{London, United Kingdom}{A.~Ricci, W.~Yeoh, N.~Agmon, B.~An (eds.)}
\copyrightyear{2023}
\acmYear{2023}
\acmDOI{}
\acmPrice{}
\acmISBN{}

\acmSubmissionID{789}

\title{Fictitious Cross-Play: Learning Global Nash Equilibrium in\\ Mixed Cooperative-Competitive Games}

\author{Zelai Xu}
\affiliation{
  \institution{Tsinghua University}
  \city{Beijing}
  \country{China}
}
\email{zelai.eecs@gmail.com}

\author{Yancheng Liang}
\affiliation{
  \institution{Tsinghua University}
  \city{Beijing}
  \country{China}
}
\email{liangyc19@mails.tsinghua.edu.cn}

\author{Chao Yu}
\affiliation{
  \institution{Tsinghua University}
  \city{Beijing}
  \country{China}
}
\email{zoeyuchao@gmail.com}

\author{Yu Wang}
\affiliation{
  \institution{Tsinghua University}
  \city{Beijing}
  \country{China}
}
\email{yu-wang@tsinghua.edu.cn}

\author{Yi Wu}
\affiliation{
  \institution{Tsinghua University}
  \country{Shanghai Qi Zhi Institute}
}
\email{jxwuyi@gmail.com}

\DeclareMathOperator*{\argmax}{arg\,max}

\newtheorem{definition}{Definition}

\begin{abstract}

Self-play (SP) is a popular multi-agent reinforcement learning (MARL) framework for solving competitive games, where each agent optimizes policy by treating others as part of the environment. Despite the empirical successes, the theoretical properties of SP-based methods are limited to two-player zero-sum games. However, for mixed cooperative-competitive games where agents on the same team need to cooperate with each other, we can show a simple counter-example where SP-based methods cannot converge to a global Nash equilibrium (NE) with high probability. Alternatively, Policy-Space Response Oracles (PSRO) is an iterative framework for learning NE, where the best responses w.r.t. previous policies are learned in each iteration. PSRO can be directly extended to mixed cooperative-competitive settings by jointly learning team best responses with all convergence properties unchanged. However, PSRO requires repeatedly training joint policies from scratch till convergence, which makes it hard to scale to complex games. In this work, we develop a novel algorithm, \emph{Fictitious Cross-Play} (FXP), which inherits the benefits from both frameworks. FXP simultaneously trains an SP-based main policy and a counter population of best response policies. The main policy is trained by fictitious self-play and cross-play against the counter population, while the counter policies are trained as the best responses to the main policy's past versions. We validate our method in matrix games and show that FXP converges to global NEs while SP methods fail. We also conduct experiments in a gridworld domain, where FXP achieves higher Elo ratings and lower exploitabilities than baselines, and a more challenging football game, where FXP defeats SOTA models with over 94\% win~rate.

\end{abstract}


\newcommand{\BibTeX}{\rm B\kern-.05em{\sc i\kern-.025em b}\kern-.08em\TeX}

\begin{document}


\pagestyle{fancy}
\fancyhead{}


\maketitle 


\section{Introduction}\label{sec:intro}

Self-play (SP) has been the most popular paradigm for multi-agent reinforcement learning (MARL), where agents collect training experiences by playing against themselves and adopt single-agent RL algorithms for policy improvement by treating other agents as part of the environment.
This framework has led to great advances in a wide range of scenarios, including fully cooperative games~\cite{bard2020hanabi}, two-player competitive games~\cite{silver2016mastering, vinyals2019grandmaster}, and even mixed cooperative-competitive games~\cite{jaderberg2019human, berner2019dota}.

Despite these empirical successes, the theoretical convergence properties of SP are limited to two-player zero-sum games, where the average policies of no-regret algorithms in SP are guaranteed to converge to a Nash equilibrium (NE)~\cite{blum2007learning}. However, other settings, particularly the mixed cooperative-competitive games, are largely unstudied. 
Existing works often directly apply the MARL methods originally designed for two-player zero-sum games to more general settings and assume strong results can be still achieved.

Unfortunately, we show a simple counter-example where SP methods converge to a suboptimal joint policy that is exploitable by an adversary team.
This is because agents in popular MARL algorithms treat both their teammates and opponents as part of the environment and optimize their own policies in a fully decentralized fashion.
As a result, the team's joint policy is likely to converge to a \emph{local} NE where no single agent can improve the return by changing its policy unilaterally, but the team can \emph{jointly} change their policies to get a higher return towards a \emph{global} NE.

To inherit the convergence properties in two-player zero-sum games and to find global NE that is unexploitable by any adversary team, agents from the same team are supposed to cooperatively optimize their joint policy in mixed cooperative-competitive games.
Policy-Space Response Oracles (PSRO)~\cite{lanctot2017unified} is an alternative framework that generalizes the double oracle (DO)~\cite{mcmahan2003planning} algorithm and is guaranteed to converged to a NE in two-player games. 
PSRO maintains a population of policies and a distribution (i.e., meta-policy) over the policy pool.
In each PSRO iteration, it trains the best response (BR) to the maintained mixed strategy according to the meta-policy and adds this BR policy as a new one to the policy pool. 
When applied to mixed cooperative-competitive games, each PSRO iteration solves a \emph{fully cooperative} game by playing against a \emph{fixed} opponent policy. Therefore, we can view each team of agents as a joint one and accordingly inherit all the convergence properties of PSRO from the two-player zero-sum setting. 
However, since PSRO requires finding a joint best response in each iteration, in order to promote exploration and avoid being trapped in a local sub-optimum, the BR policy needs to be trained from scratch in every iteration. 
This can be particularly expensive and sample inefficient in complex multi-agent games. 
In addition, PSRO may have to fully explore the entire policy space before converging to an NE, resulting in a substantial large number of iterations in practice. 
Thus, despite its theoretical properties, PSRO has been much less utilized than SP in real-world applications. 


In this work, we propose a new algorithm, Fictitious Cross-Play (FXP), for learning global NE in mixed cooperative-competitive games. FXP aims to bridge the gap of SP and PSRO by training an SP-based \emph{main policy} and a BR-based \emph{counter population} of policies.
The main policy aims to produce the final global NE and is trained by a mixed strategy over self-play, fictitious play against its past versions, and cross-play against the counter population.
The counter population aims to exploit the main policies and help them get out of local NEs by cross-play against past versions of main population.
We remark that a majority of games played by FXP has a team of policies being fixed, leading to a cooperative learning nature, which helps shape the main policy towards the global NE. Meanwhile, since the main policy is still trained by self-play, FXP is able to empirically achieve much faster policy improvement than iterative BR-based methods.

We first show in matrix games that FXP quickly converges to the global NE while SP and PSRO fail within the same amount of training steps.
Then we evaluate our algorithm on the gridworld MAgent Battle environment and achieves a much lower exploitability and a Elo rating over 200 points higher than six baselines. 
Finally, we scale up FXP to tackle the challenging  11-vs-11 multi-agent full game in the Google Research Football (GRF) \cite{kurach2020google} environment.
We compare the FXP agent with the hardest built-in AI, an imitation-learning agent~\cite{huang2021tikick}, and a PSRO-based agent~\cite{liu2021towards} and achieve higher goal differences than all baselines against reference policies of different levels.
We also let FXP play against available models including built-in hard AI and Tikick, and achieve over 94\% win rates with goal differences over 2.7. 


\begin{figure*}[t]
\centering
\includegraphics[width=0.9\linewidth]{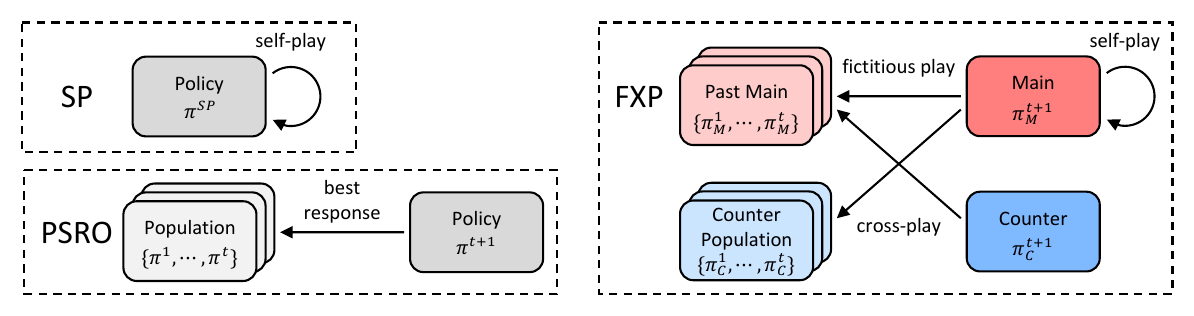}
\vspace{-4mm}
\caption{Frameworks of SP, PSRO, and FXP. SP learns a single policy against itself. 
PSRO learns a policy population by iteratively adding a best response to the current population. FXP learns a main policy and a counter population. The main policy is trained by fictitious self-play and cross-play. The counter policies are learned against past versions of main policy.} 
\label{fig:alg:fxp}
\vspace{-2mm}
\end{figure*}

\vspace{-1em}
\section{Related Work}
MARL methods have been applied to tackle a wide range of multi-agent applications~\cite{rashid2018qmix, yu2021surprising,silver2016mastering, bansal2017emergent,lowe2017multi, baker2019emergent}.
In competitive settings, self-play MARL has been proven effective in a wide range of games, from Backgammon \cite{tesauro1994td} to Go \cite{silver2016mastering} and video games \cite{vinyals2019grandmaster}. 
Fictitious self-play (FSP) \cite{heinrich2015fictitious} combines fictitious play (FP) \cite{brown1951iterative} with self-play in extensive-form games and is proved to converge to a NE in the time average.
Neural Replicator Dynamics (NeuRD) \cite{hennes2020neural} is another method with time-average convergence via self-play which approximates replicator dynamics using a policy gradients variant.
Some recent works \cite{perolat2021poincare, sokota2022unified} also achieve last-iteration Nash in two-player zero-sum games by adding regularization to Follow the Regularized Leader (FoReL) and mirror descent.

Another line of work is based on the game-theoretic algorithm double oracle (DO) \cite{mcmahan2003planning}.  
Policy-Space Response Oracles (PSRO) \cite{lanctot2017unified} is the most popular generalization of DO, which trains a population of policies by iteratively adding a best response to the opponent's Nash mixed strategy. PSRO is guaranteed to converge to a NE in two-player games.
Extensive-Form Double Oracle (XDO) \cite{mcaleer2021xdo} generalizes PSRO to extensive-form games by mixing best responses at every infostate instead of only at the root of the game.
$\alpha$-Rank PSRO \cite{muller2019generalized} replaces NE with a new solution concept $\alpha$-Rank and extends PSRO to $n$-player general-sum games.
Anytime PSRO \cite{mcaleer2022anytime} and Online Double Oracle (ODO) \cite{dinh2021online} combine PSRO with no-regret algorithms and online learning, respectively, and show faster convergence rates in some games like poker.
Some other variants \cite{balduzzi2019open, perez2021modelling, liu2021towards} incorporate different diversity metrics with PSRO, and achieve lower exploitability in games with high non-transitivity.

In mixed cooperative-competitive settings, many complex real-world games are solved by combining existing SP or PSRO approaches with large-scale training.
In the hide-and-seek game \cite{baker2019emergent}, agents show emergent behaviours like tool use by multi-agent self-play.
For-The-Win \cite{jaderberg2019human} adopts a population-based training framework and demonstrates human-level play in the Capture-the-Flag game.
OpenAI Five \cite{berner2019dota} adopts a past-sampling augmented self-play framework and defeated world the champion in Dota 2.
Google Research Football (GRF) \cite{kurach2020google} is another mixed cooperative-competitive game where very few works have shown strong performances in the 11-vs-11 full game.
Tikick \cite{huang2021tikick} trains the first learning-based agent that can take over the full game by imitation learning.
\cite{liu2021towards} uses a diversity-aware and online variant of PSRO and defeats the hardest built-in bot. We take them as our baselines. 
\label{sec:related}

\section{Preliminary}
In this section, we first establish the prerequisite definitions and notations in normal-form games,
and then describe the extension to MARL settings using empirical game-theoretic analysis.
We also formally describe the SP and PSRO algorithms.

\subsection{Normal-Formal Games}
A $K$-player \textit{normal-form game (NFG)} is often described by a tuple $(K, \Pi, U)$. 
Each player $k\in[K]$ has a finite set of pure strategies $\Pi_k = \{\pi_k^1, \cdots, \pi_k^{\|\Pi_k\|}\}$ and $\Pi = \times_{k=1}^K \Pi_k$ is the set of all pure strategy profiles (or joint strategy).
For each pure strategy profiles $\pi\in\Pi$, the utility function $U: \Pi \to \mathbb{R}^K$ gives a vector $U(\pi)=(U_1(\pi), \cdots, U_K(\pi))$ where $U_k(\pi)$ is the payoff value of player $k$ under strategy profile $\pi$.
The goal of each player is to maximize its own expected utility by choosing a pure strategy $\pi_k$ or sampling from a mixed strategy $\sigma_k \in \Delta(\Pi_k)$.

We consider the setting of \textit{mixed cooperative-competitive games}, where the $K$ players are divided into two competing teams of size $N=K/2$.
Players within the same team are fully cooperative and share the same utility.
Let $U_{i, n}, i\in\{1, 2\}, n\in[N]$ denotes the utility function of player $n$ in team $i$, we have
\begin{align}
    U_{i, 1}(\pi) &= \cdots = U_{i, N}(\pi) = U_{t_i}(\pi),\ \forall \pi \in \Pi, i \in \{1, 2\}.
\end{align}
On the other hand, the two teams are fully competitive and their utilities sum to zero, i.e.,
\begin{align}
    U_{t_1}(\pi) + U_{t_2}(\pi) = 0,\ \forall \pi \in \Pi.
\end{align}

Given a mixed strategy profile $\sigma_{-k}$ of all players other than player $k$, the \textit{best response (BR)} of player $k$ is defined as $\mathrm{BR}(\sigma_{-k}) = \arg\max_{\pi_k\in\Pi_k}\mathbb{E}_{\pi_{-k}\sim\sigma_{-k}}[U_k(\pi_k, \pi_{-k})]$.
A mixed strategy profile $\sigma$ is a \textit{Nash equilibrium (NE)} if
\begin{equation}
    \sigma_k=\mathrm{BR}(\sigma_{-k}), \ \forall k\in[K].
    \label{eq:ne}
\end{equation}
Similarly, for mixed cooperative-competitive games, we can define $\mathrm{BR_{team}}(\sigma_{t_{-i}}) = \arg\max_{\pi_{t_i}\in\Pi_{t_i}} \mathbb{E}_{\pi_{t_{-i}}\sim\sigma_{t_{-i}}} [U_{t_i}(\pi_{t_i}, \pi_{t_{-i}})]$ to be the \textit{team best response (team BR)}, where $\sigma_{t_{-i}} = (\sigma_{-i, 1}, \cdots, \sigma_{-i, N})$ is the opponent team's joint mixed strategy and $\Pi_{t_i} = \times_{n=1}^N \Pi_{i, n}$ is the set of all joint pure strategies of team $i\in\{1, 2\}$.
We use \textit{local Nash equilibrium (local NE)} to refer to a mixed strategy that satisfies Equation~(\ref{eq:ne}) in mixed cooperative-competitive games, and use \textit{global Nash equilibrium (global NE or team NE)} to refer to a mixed strategy $\sigma=(\sigma_{t_1}, \sigma_{t_2})$ such that
\begin{equation}
    \sigma_{t_i}=\mathrm{BR_{team}}(\sigma_{t_{-i}}),\ \forall i\in\{1,2\}.
    \label{eq:gne}
\end{equation}
It is worth noting that a global NE is always a local NE, but a local NE is not necessarily a global NE.
The goal of mixed cooperative-competitive games is to learn a global NE, and the metric to evaluate a mixed strategy profile $\sigma$ is \textit{team exploitability} $e_{\mathrm{team}}(\sigma)=\sum_{i\in\{1, 2\}} U_{t_{-i}}(\mathrm{BR_{team}}(\sigma_{t_i}), \sigma_{t_i})$, which can be roughly interpreted as the "distance" from $\sigma$ to a global NE.
Note that the local NE defined here is different from the term that refers to the locality in the action space of continuous games in other works like~\cite{ratliff2013characterization}.

\subsection{Extension to MARL}

A Markov game (MG)~\citep{littman1994markov} defined as a tuple $(K, \mathcal{S}, \mathcal{A}, \mathcal{O}, O, r, P, \gamma)$. 
Here, $K\in\mathbb{R}$ is the number of agents, $\mathcal{S}$ is the state space, $\mathcal{A}, \mathcal{O}$ are the action space and observation space shared across all agents, and $\gamma\in[0, 1]$ is the discount factor.
Given states $s, s'\in\mathcal{S}$ and joint action $\bm{a}\in\mathcal{A}^K$, $o_k = O_k(s)$ and $r_k(s, \bm{a})$ are the local observation and reward of agent $k$, and $P(s, \bm{a}, s')$ is the transition probability from state $s$ to $s'$ under joint action $\bm{a}$. 
Each agent uses a policy $\pi_k(a_k|o_k)$ to produce its action $a_k$ from the local observation $o_k$, and the expected return of agent $k$ under joint policy $(\pi_k, \pi_{-k})$
is $J_k(\pi_k, \pi_{-k}) = \mathbb{E}_{s^t, \bm{a}^t}[\sum_t \gamma^t r_k(s^t, \bm{a}^t)]$.
Many popular MARL algorithms like MAPPO \cite{yu2021surprising} follow the \emph{decentralized learning} framework, i.e., each agent optimizes the its return by treating other agents as part of the environment.
Given other agents' joint policy $\pi_{-k}$, these methods aim to find the optimal policy $\pi_k^*$ w.r.t.
\begin{align}
    \pi_k^* = \argmax_{\pi_k} J_k(\pi_k, \pi_{-k}).
    \label{eq:marl_obj}
\end{align}

For complex games with prohibitively large policy space, MARL is often combined with \textit{empirical game-theoretic analysis (EGTA)} to construct a higher-level normal-form game, and apply game-theoretic analysis in this meta-game to guide the learning of new policies.
In the normal-form meta-game, the pure strategies become \textit{policies} learned by MARL algorithms, the set of current policies $\Pi$ is also called a \textit{population}, and the mixed strategy $\sigma$ is called a \textit{meta-policy}.
An \textit{empirical payoff matrix} $U$ can be constructed by simulating in the original game for all joint policy combinations.
Since the population can get larger with more policies learned and is no longer fixed, we use $\mathrm{BR}(\sigma\Pi)$ to denote the BR of population $\Pi$ with meta-policy $\sigma$ and $\mathrm{BR}(\pi)$ to denote the BR of policy $\pi$.
Given a joint policy $\pi=(\pi_k, \pi_{-k})$, the utility function of agent $k$ is its expected return in the original game $U_k(\pi) = J_k(\pi_k, \pi_{-k})$, and the BR of $\Pi_{-k}$ with $\sigma_{-k}$ becomes
\begin{align}
    \mathrm{BR}(\sigma_{-k}\Pi_{-k}) = \argmax_{\pi_k}\mathbb{E}_{\pi_{-k}\sim\sigma_{-k}}[J_k(\pi_k, \pi_{-k})],
    \label{eq:br}
\end{align}
which is equivalent to Equation~(\ref{eq:marl_obj}) by sampling joint policy $\pi_{-k}$ according to the meta-policy $\sigma_{-k}$ at the beginning of each episode.
Therefore, we can use MARL algorithms as approximate BR and team BR oracles in the meta-game.


\subsection{Self-play}
Self-play learns a single policy by training against itself. 
Using RL as the approximate BR oracle, SP starts with a randomly initialized policy and repeatedly updates the policy toward the BR of itself. 
SP is simple and efficient in learning.
Fictitious Play (FP) extends SP by training a policy against its time-averaged policy $\overline{\pi}^{FP}$ rather than $\pi^{FP}$ itself, and the time-averaged policy of FP is guaranteed to converge to a NE.
The pseudocode of SP is listed in Algorithm~\ref{alg:sp}.

\begin{algorithm}[bt!]
\caption{Self-Play (SP)}
\label{alg:sp}
\textit{Input:} Randomly initialized policy $\pi^{SP}$ \\
\For{many episodes}{
    Update $\pi^{SP}$ toward $\mathrm{BR}(\pi^{SP})$ \\
}
\textit{Output:} {Policy $\pi^{SP}$}
\end{algorithm}

For mixed cooperative-competitive games, one can use MARL to find the approximate team BRs.
However, with decentralized learning, each agent optimizes its own policy rather than the team one, easily yielding a suboptimal joint policy. 
Therefore, it is very likely that the SP policy converges to a local NE where no single agent can improve unilaterally, but the team policy can still get a higher return by jointly optimizing the policies towards a global NE.
We present a concrete example with detailed analysis in Sec.~\ref{sec:example}.

\subsection{Policy-Space Response Oracles}

Instead of training a single policy, PSRO iteratively trains a population of policies to find the NE of large games.
PSRO starts with an initial population $\Pi^1 = \{\pi^1\}$ with a single random policy. 
In iteration $t$, an empirical payoff matrix $U$ is computed by simulations using policies in the current population $\Pi^t$.
The payoff matrix $U$ is then used by a meta-solver to calculate the meta-policy $\sigma$ of population $\Pi^t$, and a new policy $\pi^{t+1}$ is trained to be the BR of population $\Pi^t$ with meta-policy $\sigma$.
The new policy is added to the population and PSRO continues to the next iteration.
PSRO generalizes many algorithms by using different meta-solvers.
FP can be regarded as an instance of PSRO with uniform solver which assigns equal probability to each policy.
DO is also an instance of PSRO with Nash solver which uses the NE of the restricted game as the meta-policy.
Other meta-solvers include projected replicator dynamics (PRD) solver \cite{lanctot2017unified}, rectified Nash solver \cite{balduzzi2019open}, $\alpha$-Rank solver \cite{muller2019generalized}, etc.
The pseudocode of PSRO is listed in Algorithm~\ref{alg:psro}.

\begin{algorithm}[bt!]
\caption{Policy-Space Response Oracles (PSRO)}
\label{alg:psro}
\textit{Input:} Initial population with random policy $\Pi^1 = \{\pi^1\}$ \\
\For{$t = 1, 2, \cdots, T$}{
    Update payoff matrix $U$ by game simulations \\
    $\sigma \leftarrow$ meta-solver($U$) \\
    \For{many episodes}{
        Update $\pi^{t+1}$ toward $\mathrm{BR}(\sigma\Pi^t)$ \\
    }
    $\Pi^{t+1} \leftarrow \Pi^t \cup \{\pi^{t+1}\}$ \\
}
\textit{Output:} {Population $\Pi^{T+1}$ and meta-policy $\sigma$}
\end{algorithm}

PSRO is guaranteed to converge to a NE in two-player games with proper meta-solvers, 
and can be directly extended to mixed cooperative-competitive games by using a team BR oracle.
This is because in each iteration, the BR policy is trained against a mixture of \emph{fixed} policies yielding a fully cooperative learning problem with stationary opponents.
However, to avoid struggling in poor local sub-optimum, PSRO has to train the policy from scratch in each iteration in order to find the global best response.
In addition, PSRO may have to fully explore the policy space to cover all the strategy modes before converging to a global NE. 
Taking Rock-Paper-Scissors (RPS) as an example,
PSRO has to cover all three modes to find the NE $(1/3, 1/3, 1/3)$. 
These issues make PSRO very inefficient in complex games with a huge policy space. 
\label{sec:background}

\section{A Motivating Example}\label{sec:example}
Here we introduce an illustrative mixed cooperative-competitive game,  i.e., a normal-form game with two competitive teams of $N$ homogeneous agents. 
Each agent can choose from two actions $0$ or $1$. The utility function $U$ has $U(x, y) = -U(y, x)$ and satisfies
\begin{align*}
    &U(0_N,1_N) = C, \\
    &U(0_N, y) = \epsilon \sum_{i=1}^N y_i, & \forall &y\ne 1_N, \\
    &U(x, y) = \sum_{i=1}^N x_i-y_i, & \forall &x, y\ne 0_N. \\
\end{align*}
Here the parameters $C,\epsilon$ satisfy $0<\epsilon\ll C\ll N$. When there is no ambiguity, we use $\mathbf{0},\mathbf{1}$ to represent the joint policy that corresponding agents all act $0$ or all act $1$, respectively.
Clearly, the game has a unique global NE $(\mathbf0, \mathbf0)$, and a local suboptimal NE $(\mathbf1, \mathbf1)$.

Let the learning policy and the opponent policy be $\pi, \mu$, respectively. Thus for self-play, $\mu^t=\pi^t$, and for PSRO and our counter policy, $\mu$ is a fixed policy against which the best response is learned.

\begin{definition}[Q-function]
    At each time $t$, the Q-function 
    $
        Q_i^t(a_i) = \mathbb{E}_{\bm{x}_{-i}\sim \pi_{-i}^t,\bm{y}\sim \mu^t} U([a_i, \bm{x}_{-i}], \bm{y}) 
    $ is computed for each agent $1\le i \le N$ and action $k\in\{0,1\}$. 
\end{definition} 

\subsection{Self-play and Its Variants}

We show that under decentralized learning, typical SP-based methods no longer converge to a global NE with a mild assumption. 

\begin{definition}[Preference Preservation]
    We say a learning process is preference preservation if the relative ratio of choosing action $x$ and $y$ keeps increasing when all the past observed Q-function of $x$ is larger than $y$, and the ratio updating rules are monotone with $Q$. To be more specific,
    \begin{equation}
        \label{def:ratio-increase}
        \forall t'\le t,  Q_i^{t'}(x) \ge Q_i^{t'}(y)\Rightarrow \frac{\pi_i^{t+1}(x)}{\pi_i^{t+1}(y)} \ge \frac{\pi_i^{t}(x)}{\pi_i^{t}(y)}
    \end{equation}
    and $\forall t\ge 0, 1\le i\le N, x, y \in \Pi_i, \exists \text{ monotone non-decreasing }f_{i,x,y}^t$ such that
    \begin{align}
        \label{def:monotone-update}
        &\forall t'\le t, z\notin\{x,y\}, \pi_i^{t'}(z)=0 \\ \notag \Rightarrow& \frac{\pi_i^{t+1}(x)}{\pi_i^{t+1}(y)} = f_{i,x,y}^t\left( \frac{\pi_i^{t}(x)}{\pi_i^{t}(y)}, \{Q_x^s-Q_y^s\}_{s=0}^t \right)
    \end{align}
\end{definition}

This property holds for many SP-based algorithms, including FSP \cite{heinrich2015fictitious, heinrich2016deep}, Follow the Regularised Leader \cite{shalev2012online}, Replicator Dynamics \cite{hennes2020neural}, Multiplicative Weights Update \cite{freund1999adaptive}, Counter Factual Regret Minimization \cite{brown2019deep}, or any softmax variants of them. Although some of them are proved to converge to NE under two-player zero-sum games, we show in the following theorem that in the mixed cooperative-competitive game we proposed, none of them converge to the global NE $(\mathbf0, \mathbf0)$.

\begin{theorem}
    \label{theorem:SP-not-converge}
    Any algorithm with preference preservation will not produce a policy $\pi$ converging to the global NE if the initialized policy $\pi^0$ does satisfy
    $$
    \forall i, \pi_{-i}^0(\mathbf 0) \le \frac{1}{N+1+2C+\epsilon}.
    $$
    When the policy is randomly initialized, there is at least a probability of $1 - \exp\left(-\Omega(N) \right)$ that the above condition is satisfied and the policy does not converge to the global NE.
\end{theorem}

We list the proof in Appendix~\ref{app:proof}. The obstacle of learning towards the global NE largely comes from the partial observation, as each agent only consider its local Q-function. Despite the challenge of cooperative learning, we will show that learning against a fixed opponent rather than the varying $\pi^t$ does mitigate the problem.

\subsection{Playing Against a Fixed Opponent}

In the learning of PSRO's best response, the opponent policy $\mu$ is fixed.
Although the opponent policy can be dependent on the algorithm, our analysis is based on the opponent policy $\mu\in \{\mathbf0, \mathbf1\}$, since the game has only two local NEs $(\mathbf0, \mathbf 0), (\mathbf1, \mathbf 1)$

\begin{definition}[Good Initialization]
    A good initialization $\pi^0$ regarding a certain learning configuration enable the learned policy to converge to the global NE.
\end{definition} 

\textbf{Remark.} We omit the discussion of the existence of convergence or the convergence to other polices here, as at most cases the policy will converge to either $\mathbf 0$ or $\mathbf 1$.

Therefore, a better learning algorithm should have a larger set of good initialization. We now compare $S_\text{SP}$ (self-play) with $S_\mu$ (the fixed opponent $\mu\in \{\mathbf0, \mathbf 1\}$).

\begin{theorem}
    \label{theorem:good-initialization}
    For $\mu\in\{\mathbf0, \mathbf1\}$, when the same preference preserved algorithm is applied, we must have $S_\text{SP} \subseteq S_\mu$. And, learning against fixed $\mu$ strictly enlarges the good initialization set as $S_\mu \backslash S_\text{SP} \ne \varnothing$.
\end{theorem}

The proof is in Appendix~\ref{app:proof}.
Theorem~\ref{theorem:good-initialization} intuitively shows that cooperative learning with a fixed opponent can be much easier. 
Hence, PSRO will have a much higher chance to find a better joint policy than SP.


\section{Method}\label{sec:method}
By the motivating example, SP-based algorithms can fails to finding the global NE in mixed cooperative-competitive games because of decentralized learning. 
PSRO mitigates this issue by training against fixed opponents iteratively. 
However, PSRO can be very inefficient in complex games with a large policy space. 
Therefore, we aim to bridge the gap of SP and PSRO in this section. 

\subsection{Fictitious Cross-Play}
Fictitious Cross-Play (FXP) trains an SP-based main policy and a BR-based counter population.
The main policy aims to find the global NE of the game and is trained by fictitious self-play and cross-play against the counter population.
To prevent the main policy from local NEs, an auxiliary counter population is iteratively trained for the best responses to past versions of main policy.
The counter population is able to find better joint policies to exploit the past main policies because it is trained against fixed opponents, leading to a fully cooperative learning problem. 
The learned counter policies are then used as opponents for main policy in cross-play, which helps it get out of local NEs towards the global NE.
For ease of notations, we use main population to refer to the set of all past \emph{checkpoints} of the main policy.

FXP starts with randomly initialized policies $\pi_M^1, \pi_C^1$, and the initial main population and counter population are $\Pi_M^1=\{\pi_M^1\}, \Pi_C^1=\{\pi_C^1\}$. 
Consider the restricted game where the row player's policies are $\Pi_M$ and the column player's policies are $\Pi_C$, we denote the payoff matrix of this restricted game as $U_{M \times C}$.
Since the game is symmetric, we also have a joint population $\Pi_{M+C} = \Pi_M \cup \Pi_C$, and the corresponding payoff matrix is denoted as $U_{M+C} = U_{(M+C) \times (M+C)}$.
In each iteration, a new main policy $\pi_M^{t+1}$ and counter policy $\pi_C^{t+1}$ are trained simultaneously against different opponents.
The main policy is trained by self-play, fictitious play against the main population $\Pi_M^t$, and cross-play against the counter population $\Pi_C^t$. 
The probability of self-play is determined by a hyperparameter $\eta$, and the meta-policy $\sigma_{M+C}$ used to sample opponents from main and counter populations is computed by a meta-solver on payoff $U_{M+C}$.
Similarly, a meta-policy $\sigma_M$ for the row player in the restricted game with payoff $U_{M \times C}$ is computed, and the counter policy is train to be the best response of the main population $\Pi_M^t$ with meta-strategy $\sigma_M$.
The new main and counter policies are added to their populations after convergence or a fixed number of training steps, and the payoff matrices $U_{M+C}, U_{M \times C}$ are updated by game simulations.
The pseudocode of FXP is listed in Algorithm~\ref{alg:fxp}.

\begin{algorithm}[bt!]
\caption{Fictitious Cross-Play (FXP)}
\label{alg:fxp}
\textbf{Input:} Initial main population and counter population with random policy $\Pi_M^1 = \{\pi_M^1\}, \Pi_C^1 = \{\pi_C^1\}$ \\
\For{$t = 1, 2, \cdots, T$}{
    Update $U_{M+C}, U_{M \times C}$ by game simulations \\
    $\sigma_{M+C} \leftarrow$ meta-solver$_M$($U_{M+C}$) \\
    $\sigma_M , \sigma_C \leftarrow$ meta-solver$_C$($U_{M \times C}$) \\
    \For{many episodes}{
        Update $\pi_M^{t+1}$ toward $\mathrm{BR}(\eta \pi_M^{t+1} + (1-\eta)\sigma_{M+C}\Pi_{M+C}^t)$ \\
        Update $\pi_C^{t+1}$ toward $\mathrm{BR}(\sigma_M \Pi_M^t)$ \\
    }
    $\Pi_M^{t+1} \leftarrow \Pi_M^t \cup \{\pi_M^{t+1}\}$ \\
    $\Pi_C^{t+1} \leftarrow \Pi_C^t \cup \{\pi_C^{t+1}\}$ \\
}
\textbf{Output:} {Population $\Pi_M^{T+1}, \Pi_C^{T+1}$ and meta-policy $\sigma_{M+C}$}
\end{algorithm}


\begin{figure*}[t]
\centering
\begin{subfigure}[h]{0.19\textwidth}
    \centering
    \includegraphics[width=\linewidth,trim={2cm 0 2cm 1cm},clip]{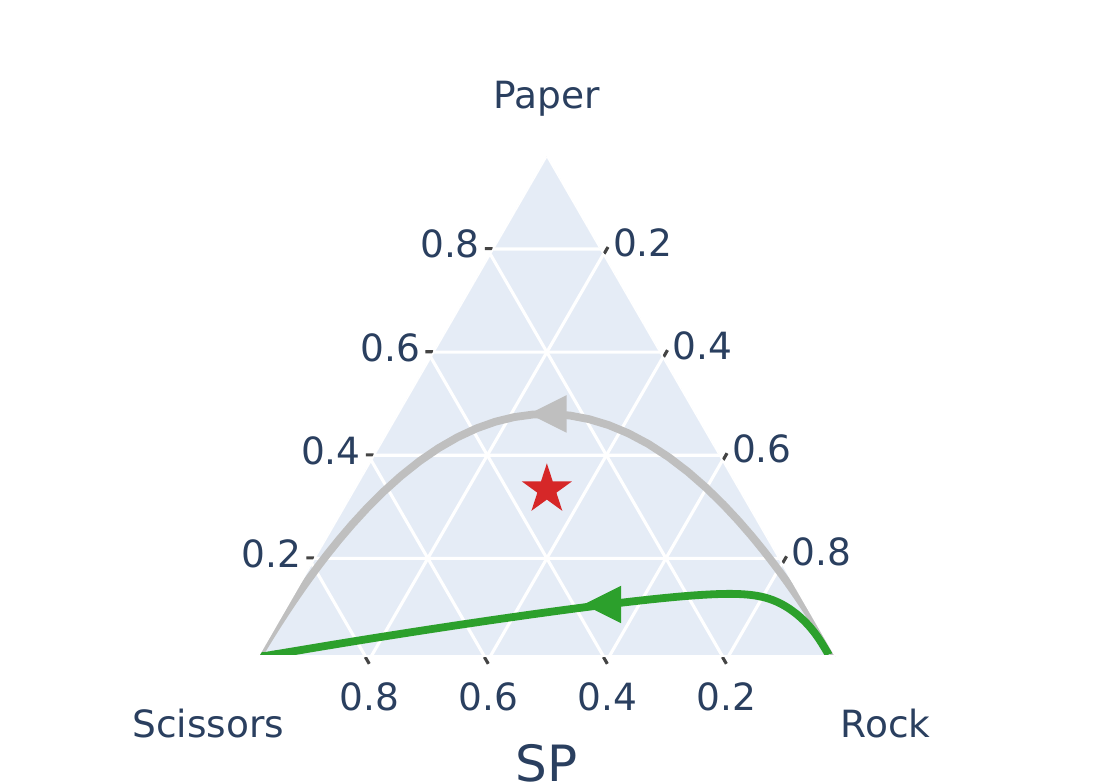}
\end{subfigure}
\hfill
\begin{subfigure}[h]{0.19\textwidth}
    \centering
    \includegraphics[width=\linewidth,trim={2cm 0 2cm 1cm},clip]{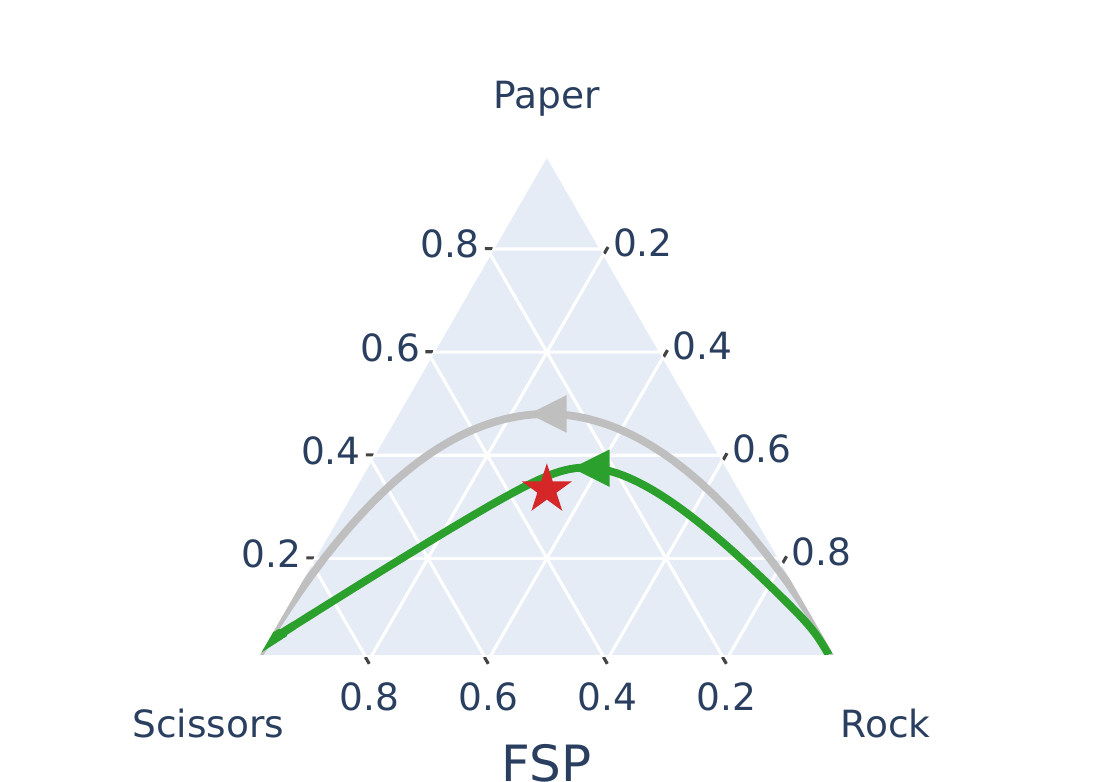}
\end{subfigure}
\hfill
\begin{subfigure}[h]{0.19\textwidth}
    \centering
    \includegraphics[width=\linewidth,trim={2cm 0 2cm 1cm},clip]{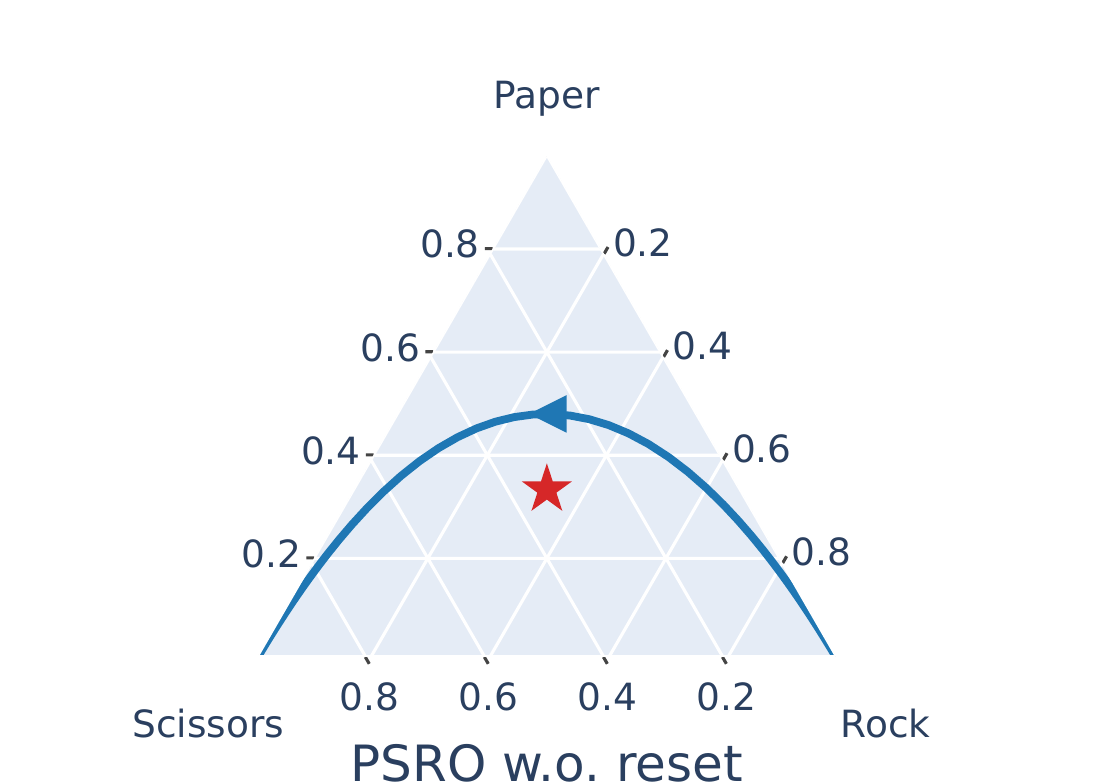}
\end{subfigure}
\hfill
\begin{subfigure}[h]{0.19\textwidth}
    \centering
    \includegraphics[width=\linewidth,trim={2cm 0 2cm 1cm},clip]{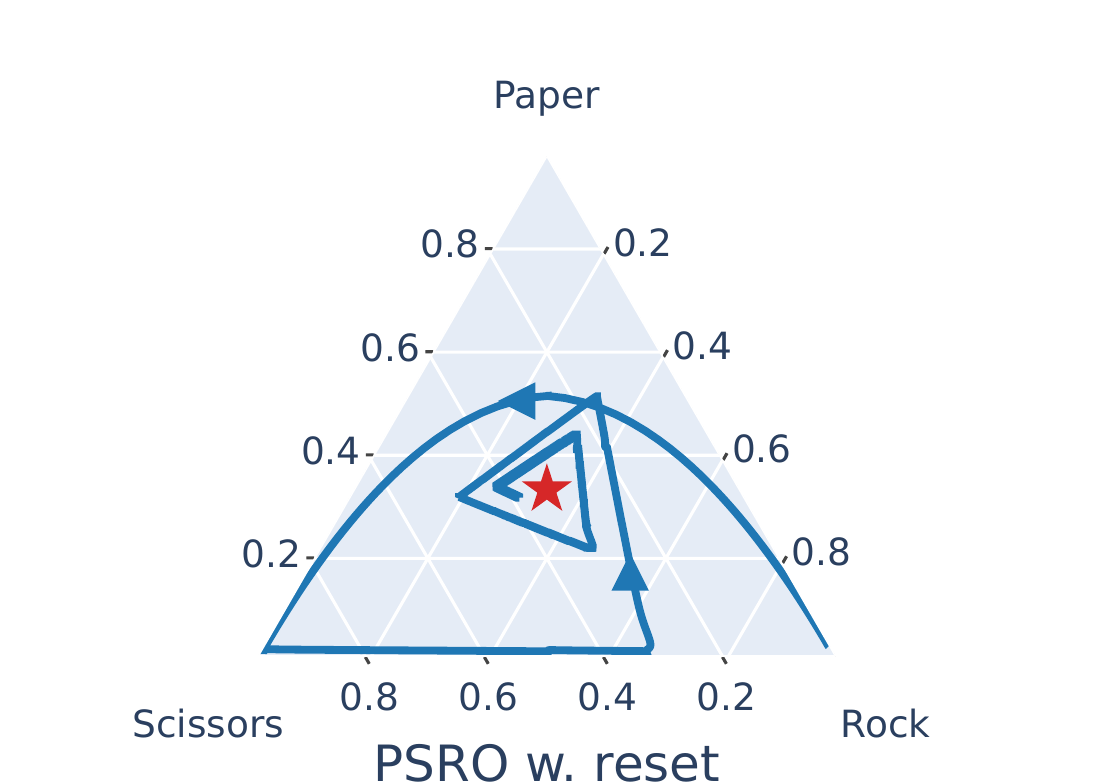}
\end{subfigure}
\hfill
\begin{subfigure}[h]{0.19\textwidth}
    \centering
    \includegraphics[width=\linewidth,trim={2cm 0 2cm 1cm},clip]{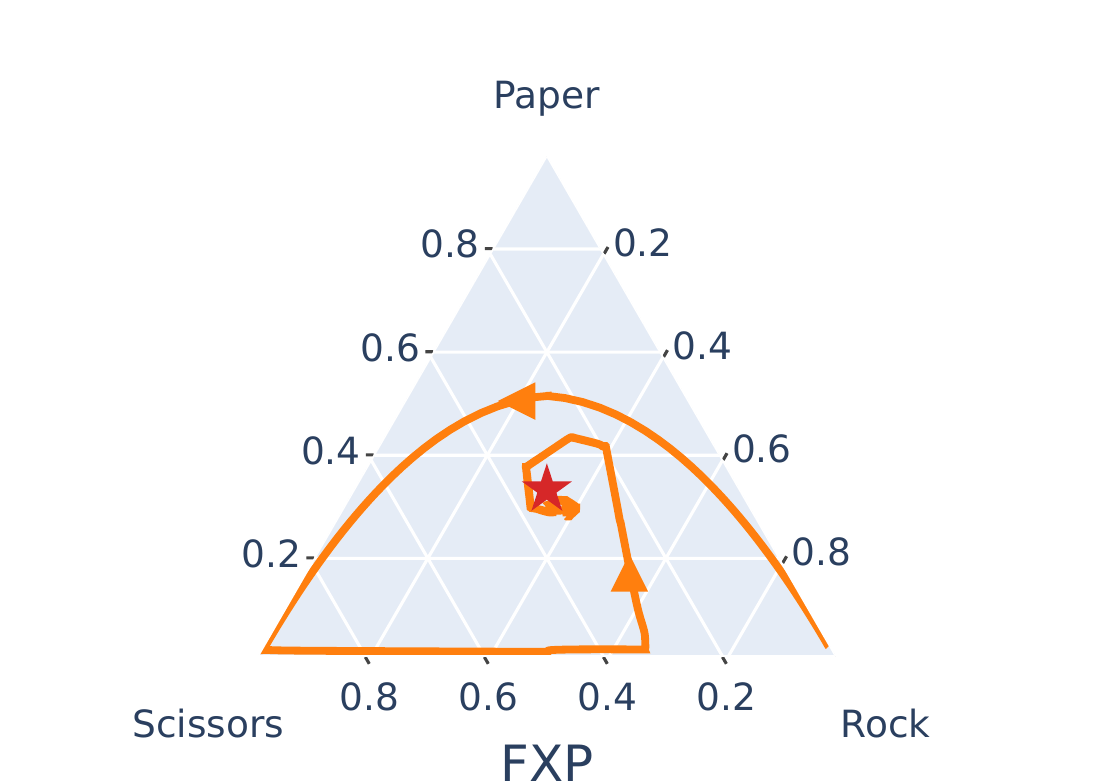}
\end{subfigure}
\vspace{-2mm}
\caption{Learning dynamics of SP, FSP, PSRO without and with reset (i.e., train from scratch), and FXP in the team RPS game. FXP quickly converges to the global NE (red star). Each algorithm is trained for the same number of steps. We counts the steps for both main and counter policies in FXP for a fair comparison.}
\label{fig:exp:trps}
\vspace{-1mm}
\end{figure*}

\subsection{Practical Implementation}

For large real-world games, we combine FXP with neural networks and use a popular MARL method, such as MAPPO~\cite{yu2021surprising} as the approximate BR oracle.
In iteration $t$, we run the current main policy and counter policy against different opponents to collect training samples.
When an episode starts, the opponent for main policy is set to itself with probability $\eta$, otherwise is sampled from the joint population $\Pi_{M+C}^t$ according to meta-policy $\sigma_{M+C}$.
Similarly, the opponent for counter policy is sampled from the main population $\Pi_M^t$ according to meta-policy $\sigma_M$.
The main and counter policies are then updated using MARL algorithms based on these samples. 
This procedure is repeated for many episodes until convergence or a maximum number of steps. 
Then the policies are added to the main and counter population to continue to the next FXP iteration. 

To accelerate training in complex games, we initialize the main policy $\pi_M^{t+1}$ in iteration $t+1$ using policy $\pi_M^t$ from the previous iteration.
This is much more efficient than training from scratch, since the current main policy is already a best response to most of the new target opponents.
On the other hand, the counter policy in each iteration remains to be trained from scratch or from an \emph{unconverged} early checkpoint. 
This is to avoid the situation where both main and counter policies are trapped in the same local sub-optimum and fail to find an approximate best response.

In practice, when the population size is large, solving meta-policies can be computationally expensive for commonly used meta-solvers. 
For efficient training, we use prioritized sampling which assigns a score to each opponents and samples them with probabilities proportional to their scores.
For main policy, we use the opponents' win rates as their scores
\begin{equation}
    s_{\pi_M}(\pi) = P\left(\pi\ \mathrm{wins}\ \pi_M\right),
\end{equation}
which makes the main policy focus on the hardest opponents and try to overcome them.
For counter policy, since it is learned from scratch or from an early checkpoint, we set the opponents' scores to be the product of their win rate and lose rate
\begin{equation}
    s_{\pi_C}(\pi) = P\left(\pi\ \mathrm{wins}\ \pi_C\right) \cdot P\left(\pi_C\ \mathrm{wins}\ \pi\right),
\end{equation}
which favors policies of about the same level as the counter policy and forms a curriculum to learn from easy to hard.

\subsection{Connections to SP and PSRO}

FXP can be regarded as an extension of both SP and PSRO with the hyperparameter $\eta$ used as a trade-off between efficiency and convergence.
If we set $\eta=1$, the main policy becomes a pure self-play policy and has no interaction with its past versions or the counter population. 
The counter policy will become the BR of the time average of the SP policy with a uniform meta-solver. 
If we set $\eta=0$, both main and counter policies are trained against fixed opponents, which is conceptually similar to PSRO.
However, even when $\eta=0$, FXP is different from PSRO in two ways.
First, FXP's meta-policies in each iteration are adaptive by prioritized sampling, while the meta-policy of PSRO is fixed. 
Second, the main policy of FXP is trained continuously and never reset, i.e., restart training from scratch, while the new policy in each PSRO iteration is reset to a random policy and trained from scratch. 
Note that it is possible to turn off reset in PSRO by warmstarting a new policy from previous ones.
However, PSRO requires a \emph{global} best response policy. 
Learning best responses with warmstart may easily get trapped in a local sub-optimum or a local NE and fail to sufficiently explore the policy space. 
We empirically find setting $\eta=0.2$ works well in many environments and use its as the default value in FXP.

\section{Experiment}\label{sec:expr}
In this section, we demonstrate the effectiveness of FXP in various mixed cooperative-competitive games.
We first study matrix games, where the payoff and team exploitability can be calculated exactly.  FXP converges to the global NE while other methods fail or use much more training steps.
Then we use MAPPO~\cite{yu2021surprising} as an approximate BR oracle and consider a gridworld environment MAgent Battle~\cite{zheng2018magent}. 
FXP achieves a lower team exploitability and a higher Elo rating  than other MARL baselines for NE.
Finally, with large-scale training, we use FXP to solve the challenging 11-vs-11 multi-agent full game in Google Research Football (GRF)~\cite{kurach2020google}.
We compare our methods with SOTA models including the hardest built-in AI, PSRO w. BD\&RD~\cite{liu2021towards} agent, and Tikick agent~\cite{huang2021tikick}. FXP achieves over 94\% win rate against available models with a significant goal difference.
Experiments on the motivating example, more ablation studies, and training details can be found in Appendix~\ref{app:detail}.


\subsection{Matrix Games}

We introduce two mixed cooperative-competitive matrix games to visualize the learning dynamics of FXP, SP, PSRO and their variants and compare their performance.

\textit{Team Rock-Paper-Scissors (team RPS) game.}
This game extends the classic 2-player zero-sum game Rock-Paper-Scissors (RPS) to a 4-player team competitive setting.
The 4 players are divided into 2 teams and play RPS between the teams.
Each player can choose either action 0 or action 1.
If both players in the same team choose action 0, then the team plays Rock; 
if both choose 1, the team plays Paper;
otherwise, the team plays Scissors.
Clearly, this game has a global NE where the team chooses Rock, Paper, Scissors with equal probability.
It also has a local NE where both players in the team choose action 1 and the team always plays Scissors.
This is because when all players other than self choose action 1, choosing action 0 would make the team play Paper, which is exploited by the opposing team's move Scissors.
However, the 2 players can jointly change their actions from 1 to 0 to play Rock and exploit the Scissors.

We run SP, FSP~\cite{heinrich2015fictitious}, PSRO$_\mathrm{Uniform}$~\cite{lanctot2017unified}, and FXP with uniform meta-solvers on the team RPS game and use policy gradient to optimize the policy for a same number of steps. 
The step count of FXP includes both main and counter policies for a fair comparison.
The learning dynamics of each algorithm is shown in Figure~\ref{fig:exp:trps}.
The red star in each subfigure is the global NE of team RPS game, the grey lines in SP and FSP subfigures are the traces of the training policies and the green lines are the traces of their time-averaged policies, the colored line in PSRO and FXP subfigures are the mixed policies of current populations.
As shown in the figure, SP and FSP converge to the local NE of Scissors and get stuck there forever, PSRO cycles around the global NE and slowly converges to it, and FXP quickly converges to the global NE.
We also run PSRO without reset on the game and it converges to the local NE as SP does.
This shows that PSRO has to train policy from scratch in each iteration to avoid struggling in local NEs.

\begin{figure*}
     \begin{subfigure}[b]{0.32\textwidth}
         \centering
         \includegraphics[width=0.9\textwidth]{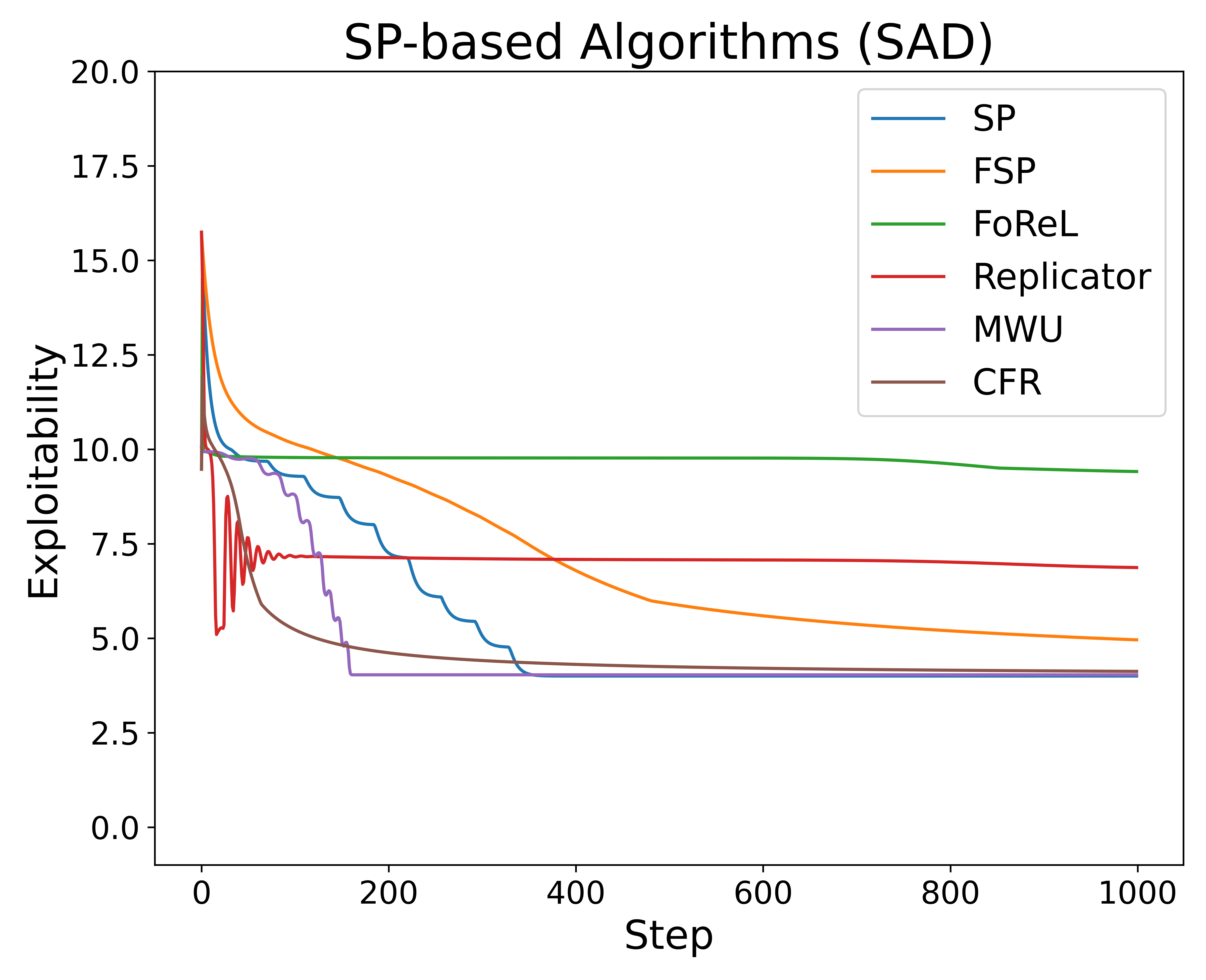}
         \vspace{-2mm}
         \caption{None of SP-based algorithms converge to the global NE that has zero exploitability.}
         \label{fig:exp:sad-sp}
     \end{subfigure}
     \hfill
     \begin{subfigure}[b]{0.32\textwidth}
         \centering
         \includegraphics[width=0.9\textwidth]{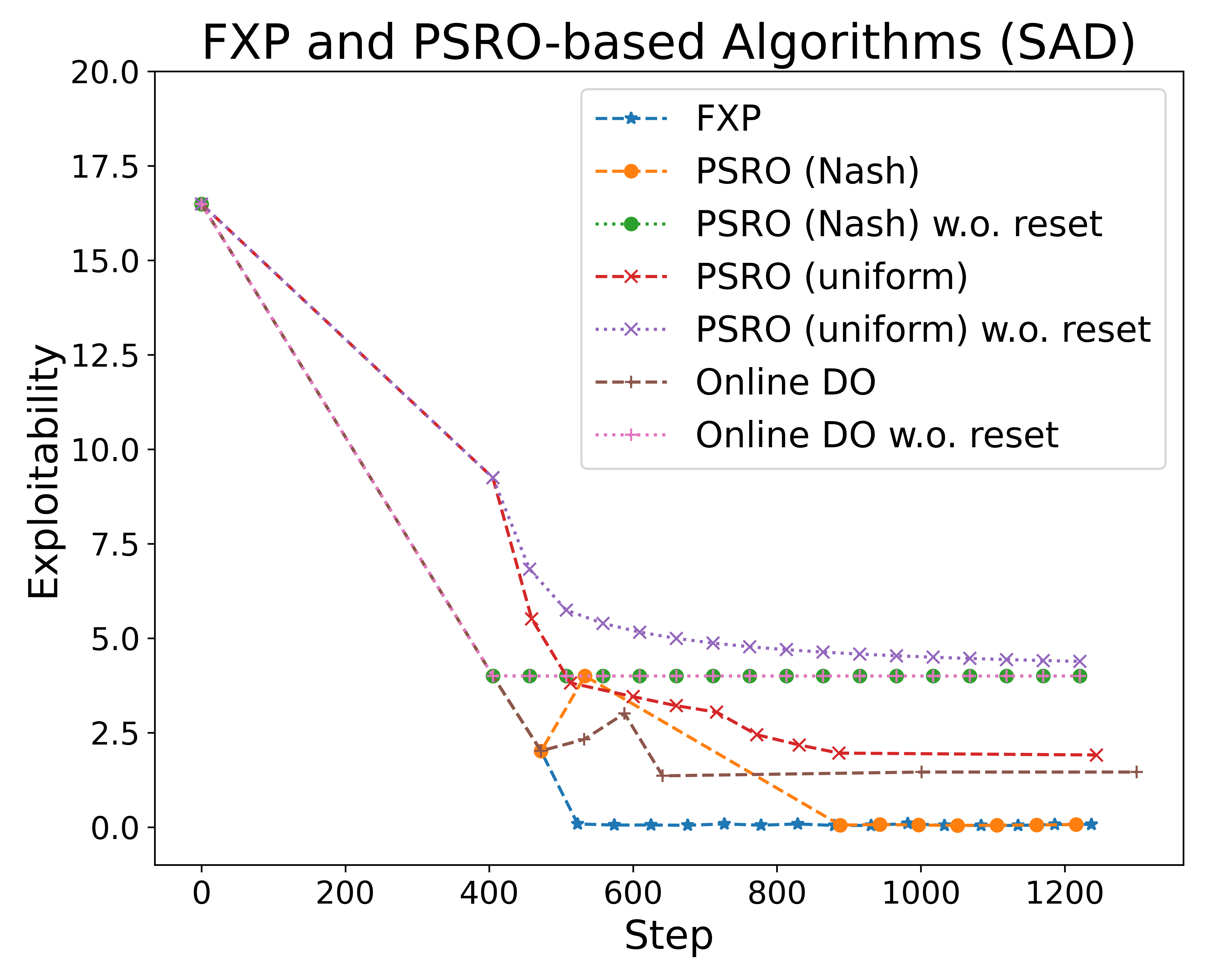}
         \vspace{-2mm}
         \caption{Exploitability is computed on the meta policy. In \textit{SAD} games FXP uses NE meta-solver.}
         \label{fig:exp:sad-psro}
     \end{subfigure}
     \hfill
    \label{fig:exp:sad}
     \begin{subfigure}[b]{0.32\textwidth}
         \centering
         \includegraphics[width=0.9\textwidth]{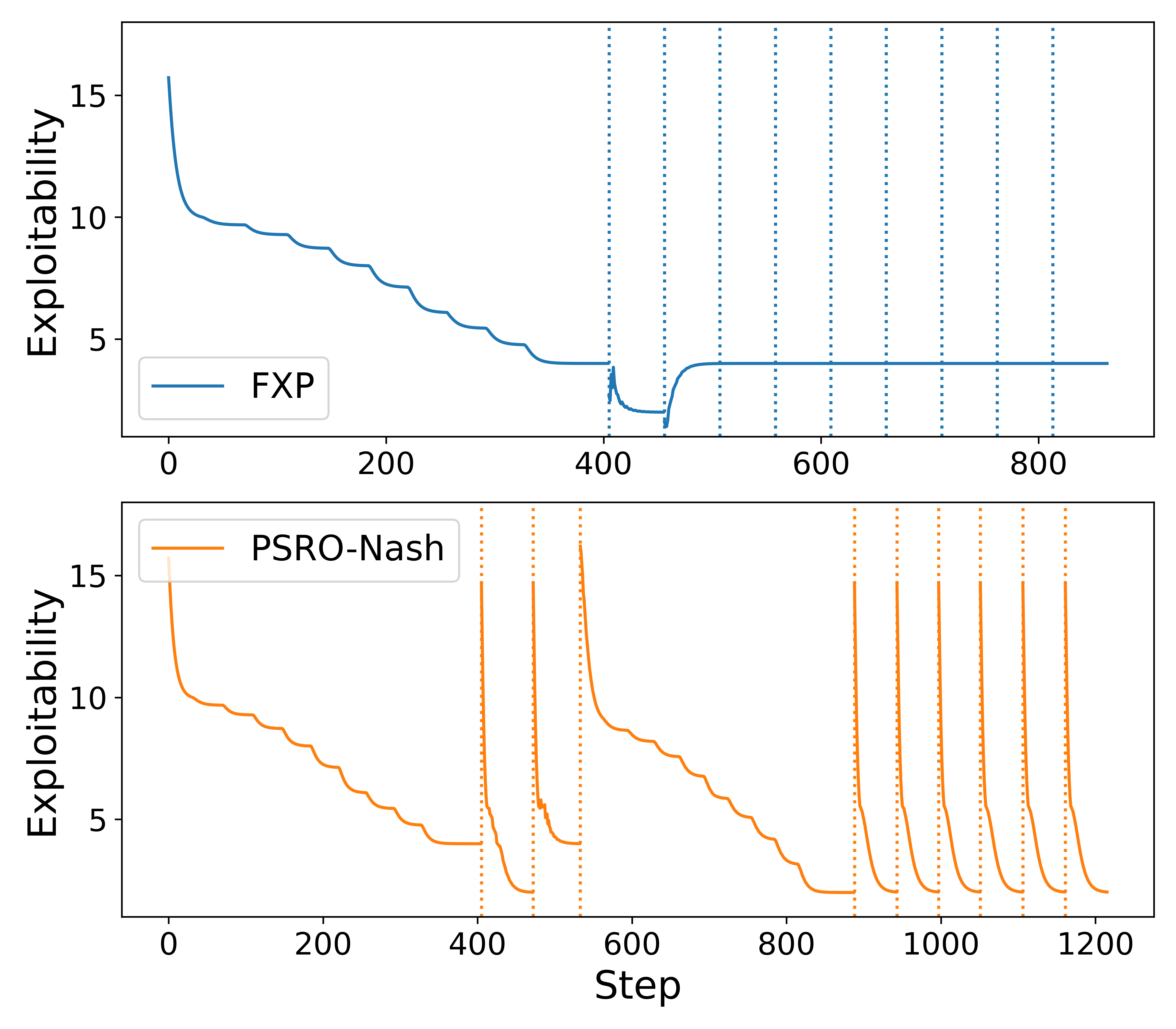}
         \vspace{-4mm}
         \caption{A vertical line means a new iteration.}
         \label{fig:exp:sad-fxp-psro}
     \end{subfigure}
     \vspace{-2mm}
    \caption{Results on seek-attack-defend (SAD) games. Evaluation metric is exploitability, which is defined as the sum of (non-negative) improvement of replacing the current policy with three following strategies (three supports of the global NE): (1) all seeking $A$; (2) 2 \textit{attack} + $(N-2)$ seeking $A$; and (3) 1 \textit{defend} + $(N-1)$ seeking A. Each step is either computing a best response or updating the  Q-function, depending on the algorithm to be used.}
    \vspace{-1mm}
\end{figure*}

\textit{Seek-attack-defend (SAD) game.} Now we propose a matrix game with a larger action space so that we can quantitatively compare different methods. A seek-attack-defend (SAD) game consists of two teams of $N$ agents, each with the action space containing $A+1$ seeking action $\{0,1,2,...,A\}$ and two special actions $\{\textit{attack},\textit{defend}\}.$ Each team seeks to obtain as much total reward as possible by cooperatively choosing seeking action $\{0,1,2,...,A\}$. A reward-level $L$ is defined as the minimum seeking action if all seeking actions differ by at most one. Otherwise, the reward-level $L$ is equal to zero. After that, the total reward $R$ is aggregated by all $R_x$ of seeking action $x$ s.t. $L\le x\le L + 1$. Therefore, teammates must learn to perform the same seeking action to receive the reward, and seek towards $A$ as reward $R_x$ gets higher as $x$ increases ($R_0=0, R_i<R_{i+1}$).

Besides reward obtaining, the team must guard their rewards. If two agents of the other team use \textit{attack} action and none of the teammates \textit{defend} the reward, the team will lose all its reward. The final utility of SAD game is defined as the difference of the reward after attack and defense are considered. Therefore, each team must properly designate some agents to attack and defend while letting others seek the highest reward $R_A$.

Here we show the learning curve of exploitability of five SP-based algorithms, including self-play (SP), fictitious self-play (FSP), follow the regularized leader (FoReL) \cite{shalev2012online}, Replicator Dynamics \cite{hennes2020neural}, multiplicative weights update (MWU) \cite{freund1999adaptive}, counter factual regret minimization (CFR) \cite{brown2019deep}. Although some of them are guaranteed to converge to NE in two-player zero-sum games, none of them converge to the global NE in SAD game, as shown in Figure~\ref{fig:exp:sad-sp}. The reason behind that is the existence of a local NE that all teammates seek with the highest action $A$, and SP-based algorithms almost always get trapped in this local NE.

Despite SP's poor performance, FXP and PSRO provide better solutions. We compare FXO with $\text{PSRO}_\text{Uniform}$ and $\text{PSRO}_\text{Nash}$.  The results in Figure~\ref{fig:exp:sad-psro} show that both FXP and $\text{PSRO}_\text{Nash}$ converge to global NE, and FXP consumes much smaller steps. (The training steps of FXP contain the cost of training counter policies for a fair comparison.) The warm-start versions of PSRO do not re-initialize the policy at the beginning of each iteration and thus degenerate to similar performance of SP.

The exploitability curves of (main) policies (NOT meta policies) in Figure~\ref{fig:exp:sad-fxp-psro} explain the advantage of FXP upon PSRO. FXP can utilize the knowledge of former policies and continue to get updated from the last iteration, while PSRO must learn skills (e.g., the cooperation of choosing the same seeking action) from scratch at each iteration. This advantage can be amplified more in larger-scale game where computing even one RL best response is non-trivial.

\subsection{MAgent Battle}

MAgent Battle is a gridworld game where a red team of $N$ agents fight against a blue team.
At each step, agents can move to one of the 12 nearest grids or attack one of the 8 surrounding grids of themselves.
Each agent has a maximum hp of 10, and lose 2 hp if is attacked by an opponent agent, and slowly recover 0.1 hp at the end of each step.
An agent is killed if its hp goes to zero and will not respawn.
The game terminates if all agents in the same team are killed or reaches a maximum number of steps.
Agents in the same team get a reward of 0.1 or 10 if an opponent agent is attacked or killed, respectively.
To make the game zero-sum between teams, agents are also penalized by 0.1 and 10 if an teammate or themselves are attacked or killed.
A good strategy in this game is to cooperatively attack the same opponent with teammates and kill opponents one by one to build an advantage in the number of agents alive.

We run SP, FSP, Neural Replicator Dynamics (NeuRD)~\cite{hennes2020neural}, PSRO$_\mathrm{Nash}$, PSRO$_\mathrm{Uniform}$, Online Double Oracle (ODO)~\cite{dinh2021online}, and FXP with MAPPO in the 3-vs-3 MAgent Battle game. 
Since the exploitability can not be exactly calculated in this game, 
we estimates the approximate exploitability of the final policies or population of different algorithms by training approximate BRs against them.
We also use Elo ratings~\cite{elo1978rating} to evaluate the relative strength of different agents.
The averaged results over 3 seeds are shown in Table~\ref{tab:exp:magent}.
Notably, FXP agents achieve the lowest exploitability and the highest Elo rating. 

\begin{figure}[bt!]
\centering
\includegraphics[width=\linewidth]{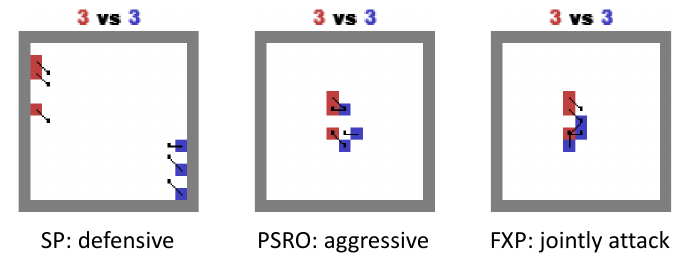}
\vspace{-5mm}
\caption{Visualization of learned behaviours by different methods in MAgent Battle. FXP learns an approximate global NE, i.e., wait for the chance to jointly attack.} 
\vspace{-2mm}
\label{fig:exp:magent}
\end{figure}

We also visualize the behaviours of agents trained by different algorithms in Figure~\ref{fig:exp:magent}.
SP converge to a defensive policy which agent stays at the edge of the map and keeps attacking in the direction of opponents, but never move toward the opponents.
This is a local NE because if only one agent tries to move and attack the opponents, it will face a dangerous 1-vs-3 situation and easily get killed.
However, it is still possible to defeat the opponents by cooperatively attacking them with all teammates.
On the other hand, PSRO agents are more aggressive because they always try to exploit a fixed population and usually overfit to a specific attacking way.
An global NE can be find if all possible attacking strategies are enumerated.
However, even in this simple gridworld game, the policy space is enormous, making PSRO methods very inefficient.
FXP agents learn an approximate global NE that is to wait and jointly attack.
This policy exploits aggressive opponents by waiting and attacking first when the opponents are trying to get close enough to them.
When facing defensive opponents, FXP agents sometimes wait forever till a tie, sometimes wait and then take the initiative to jointly attack the opponents.

\begin{table}[t]
\begin{tabular}{lcc}
    \toprule
    & Exploitability & Elo rating \\
    \midrule
    SP  & 28.66 (0.80) & 782 \\
    FSP & 21.21 (1.87) & 1627 \\
    NeuRD & 26.72 (1.43) & 1143 \\
    PSRO$_\mathrm{Uniform}$ & 24.63 (3.35) & 1495 \\
    PSRO$_\mathrm{Nash}$ & 22.54 (1.65) & 1544 \\
    ODO & 21.76 (2.19) & 1589 \\
    \midrule
    FXP & \textbf{10.62 (2.73)} & \textbf{1832} \\
    \bottomrule 
\end{tabular}
\caption{Exploitability and Elo rating of FXP agents and other MARL methods for NE in MAgent Battle game.}
\label{tab:exp:magent}
\vspace{-2em}
\end{table}

\subsection{Google Research Football}

Google Research Football (GRF) is a physics-based simulation environment adapted from popular football video games.
Each agent controls a player in the game and has to learn how to dribble the ball, cooperate with teammates to pass the ball and overcome the opponents' defense to score goals.
We consider the GRF 11-vs-11 full game, which simulates a 3000-step complete football game with standard rules. 
The long-time horizon, enormous policy spaces, and mixed cooperative-competitive nature make it a challenging problem for MARL algorithms.
We use FXP with MAPPO to solve this problem and compare with existing SOTA models.

\begin{figure}
\centering
\includegraphics[width=0.8\linewidth]{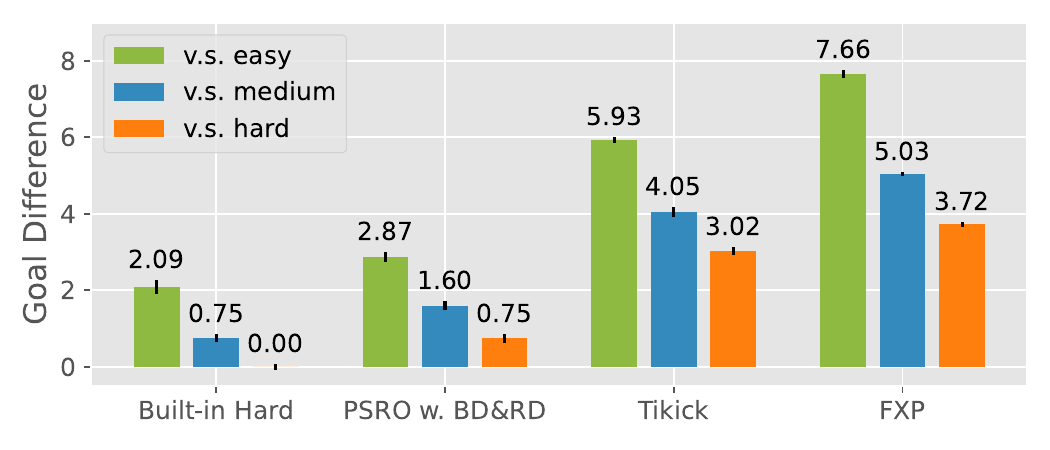}
\vspace{-1em}
\caption{Goal differences of FXP and other models against built-in AI of different levels.}
\vspace{-1em}
\label{fig:exp:fb_vs_builtin}
\end{figure}

Because the game is too complex, it is impossible to exactly calculate or approximately estimate the exploitability of a policy or a population.
As an alternative approach, we evaluate FXP and other models by playing against a set of unseen reference policies and compare their performance.
We use GRF's built-in models with different levels as the reference policies and compare FXP with SOTA models including the hardest built-in AI, a PSRO-based agent, \emph{PSRO w. BD\&RD}~\cite{liu2021towards}, an imitation learning agent \emph{Tikick}~\cite{huang2021tikick}.
Note that since the \emph{PSRO w. BD\&RD}~\cite{liu2021towards} never release their code or model. We directly report the original numbers in their paper. The model of \emph{Tikick} is released and our evaluation result of \emph{Tikick} is consistent with the paper~\cite{huang2021tikick}.
The results are shown in Figure~\ref{fig:exp:fb_vs_builtin}, where FXP achieves the largest goal difference against all reference policies.
As a reference, GRF~\cite{kurach2020google} also reports the performances of the BR policies by directly training against different level build-in AI. The BR policies achieve the average goal differences of 12.83, 5.54, 3.15 for easy, medium, hard respectively. 
We remark that, although our method has never seen the built-in models during training, FXP achieves a comparable results to BR policies, especially against medium and hard opponents. 

\begin{figure}
\centering
\hfill
\begin{subfigure}{0.4\linewidth}
    \centering
    \includegraphics[width=\linewidth]{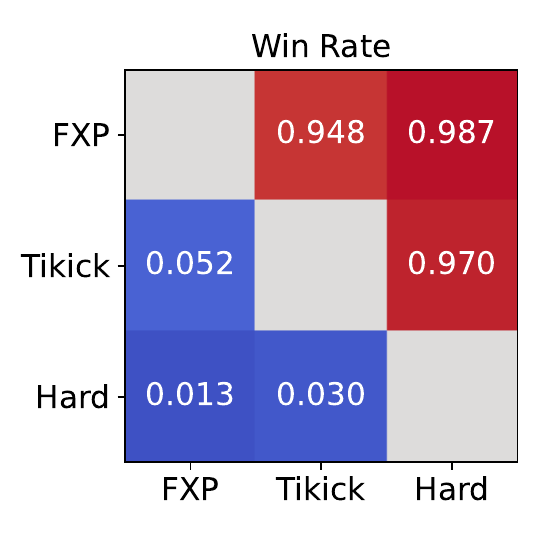}
\end{subfigure}
\hfill
\begin{subfigure}{0.4\linewidth}
    \centering
    \includegraphics[width=\linewidth]{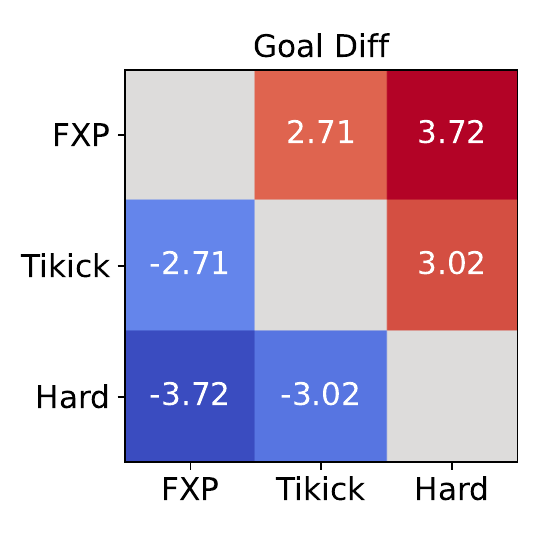}
\end{subfigure}
\hfill
\vspace{-1em}
\caption{Head-to-head win rate evaluation between FXP, Tikick and built-in hard AI in 11-vs-11 full game.}
\vspace{-1em}
\label{fig:exp:fb_h2h}
\end{figure}

Moreover,  football is a non-transitive game like RPS, so good performance against certain opponents does not necessarily means a strong policy. We also carry out a tournament-style head-to-head evaluation between FXP and available models, including Tikick and built-in hard AI. 
The results are shown in Figure~\ref{fig:exp:fb_h2h}, where FXP achieves a dominating performance, with over 94\% win rate and at least 2.7 more goals scored per game on average. 
We remark that the SOTA model Tikick performs both imitation learning on additional offline data and RL fine-tuning while FXP only adopts pure full RL training, which suggests the effectiveness of our algorithm. 

\section{Conclusion}\label{sec:conclude}
In this work, we present a novel  algorithm, Fictitious Cross-Play (FXP), to learn global NEs in mixed cooperative-competitive games.
FXP trains an SP-based main policy for the global NE and mitigates the issue of getting stuck at local NEs by training a BR-based counter population to continuously exploit the main policy.
Experiments in matrix games and gridworld games demonstrate that FXP converges to the global NE quickly and outperforms a series popular methods for NE.
FXP also defeats the SOTA models in the Google Research Football environment with a dominant win rates.
We hope FXP could bring useful insights to the community towards more effective MARL algorithms.

\begin{acks}
This research was supported by National Natural Science Foundation of China (No.U19B2019, 62203257, M-0248), Tsinghua University Initiative Scientific Research Program, Tsinghua-Meituan Joint Institute for Digital Life, Beijing National Research Center for Information Science, Technology (BNRist), Beijing Innovation Center for Future Chips and 2030 Innovation Megaprojects of China (Programme on New Generation Artificial Intelligence) Grant No. 2021AAA0150000.
\end{acks}



\bibliographystyle{ACM-Reference-Format}
\balance
\bibliography{reference}

\clearpage
\nobalance
\appendix

\section{Full Proof of Theorems}\label{app:proof}
\begin{lemma}
    For the motivating example, we first calculate the Q-funtions for policy $\pi$ and opponent $\mu$. Here we use the notation $\| \pi \|_1=\sum_{i=1}^N \pi_i(1)$. Thus,
    \begin{align}
        Q_i(0) =& \mathbb{E}_{\mathbf{x}_{-i}\sim \pi_{-i}, \mathbf{y} \sim \mu} \left[ U([0, \mathbf{x}_{-i}], y) \right] \\ \notag
        =& \mathbb{E}_{\mathbf{x}_{-i} \sim \pi_{-i}, \mathbf{y} \sim \mu} \left[  \|\mathbf{x}_{-i}\|_1 - \|\mathbf{y}\|_1 + [\mathbf{x}_{-i}=\mathbf{0} ](1+\epsilon)\|\mathbf{y}\|_1 \right. \\ \notag 
        -& [\mathbf{y}=\mathbf{0} ](1+\epsilon)\|\mathbf{x}_{-i}\|_1  \\ \notag
        +& \left. (C-N\epsilon)[\mathbf{x}_{-i}=\mathbf{0} ][\mathbf{y}=\mathbf{1} ] \right]\\ \notag
        =& \|\pi_{-i}\|_1 - \|\mu\|_1 + \pi_{-i}(\mathbf0)\|\mu\|_1(1 + \epsilon) \\ \notag
        -& \mu(\mathbf0)\|\pi_{-i}\|_1(1+\epsilon) + \pi_{-i}(\mathbf0)\mu(\mathbf1)(C - N\epsilon)
    \end{align}
    and
    \begin{align}
        Q_i(1) =& \mathbb{E}_{\mathbf{x}_{-i}\sim \pi_{-i}, \mathbf{y} \sim \mu} \left[ U([1, \mathbf{x}_{-i}], y) \right] \\ \notag
        =& \mathbb{E}_{\mathbf{x}_{-i} \sim \pi_{-i}, \mathbf{y} \sim \mu} \left[ 1 + \|\mathbf{x}_{-i}\|_1 - \|\mathbf{y}\|_1 \right. \\ \notag 
        -& [\mathbf{y}=\mathbf{0} ](1+\epsilon)(1+\|\mathbf{x}_{-i}\|_1)  \\ \notag
        -& \left. (C-N\epsilon)[\mathbf{x}_{-i}=\mathbf{1} ][\mathbf{y}=\mathbf{0} ] \right]\\ \notag
        =& 1 + \|\pi_{-i}\|_1 - \|\mu\|_1
        - \mu(\mathbf0)(1+\|\pi_{-i}\|_1)(1+\epsilon) \\ \notag
        +& \pi_{-i}(\mathbf1)\mu(\mathbf0)(C - N\epsilon)
    \end{align}
    and further,
    \begin{align}
        \label{lemma:deltaQ}
        Q_i(0)-Q_i(1) =& \mu(\mathbf0)(1+\epsilon) + \pi_{-i}(\mathbf0)\|\mu\|_1(1+\epsilon) \\ \notag
        +& (\pi_{-i}(\mathbf0)\mu(\mathbf1) + \pi_{-i}(\mathbf1)\mu(\mathbf0))(C - N\epsilon) - 1.
    \end{align}
\end{lemma}

\textbf{Proof for Theorem~\ref{theorem:SP-not-converge}}. \textit{
    Any algorithm with preference preservation does not produce a policy $\pi$ converging to the global NE if the initialized policy $\pi^0$ satisfying
    $$
    \forall i, \pi_{-i}^0(\mathbf 0) \le \frac{1}{N+1+2C+\epsilon}.
    $$
    When the policy is randomly initialized, there is at least a probability of $1 - \exp\left(-\Omega(N) \right)$ that the above condition is satisfied and the policy does not converge to the global NE.
} 

\begin{proof}
    We first show that $\forall i,t$, $$\pi_{-i}^t(\mathbf 0) \le \frac{1}{N+1+2C+\epsilon} \Rightarrow Q_i^t(0) \le Q_i^t(1).$$ Actually with $
    \mu^t=\pi^t$ and Equation~\ref{lemma:deltaQ}, we have
    \begin{align*}
        Q_i^t(0) - Q_i^t(1) \le& (1+N)(1+\epsilon)\pi_{-i}^t(\mathbf0) + 2\pi_{-i}^t(\mathbf0)(C-N\epsilon) - 1 \\
        \le& (N + 1 + 2C + \epsilon)\pi_{-i}^t(\mathbf0) - 1 \\
        \le& 0.
    \end{align*}
    Thus with ratio increase rule~\ref{def:ratio-increase}, we may derive that
    \begin{equation}
        \left. \begin{aligned}
        \pi_{-i}^t(\mathbf 0) \le \frac{1}{N+1+2C+\epsilon} \\
        \forall t'<t, Q_i^{t'}(0) \le Q_i^{t'}(1) 
        \end{aligned} \right\} \Rightarrow \pi_{-i}^{t+1} (\mathbf 0) \le \pi_{-i}^{t} (\mathbf 0).
    \end{equation}
    Therefore with induction, $\forall t$, $\pi_{-i}^t(\mathbf 0) \le \frac{1}{N+1+2C+\epsilon}$, and hence $\pi$ cannot converge to the global optimal action $\mathbf0$.
    
    Now we can see that if $\pi_j^0(0)$ is i.i.d. sampled from $[0,1]$, by Chernoff bound,
    $$
    \frac{1}{N-1}\sum_{j\ne i}\pi_j^0(0) \ge (N+1+2C+\epsilon)^{\frac{1}{N-1}}
    $$
    with probability $e^{-\Omega(N)}$ for large enough $N$ as $\lim_{N\to+\infty} (N+1+2C+\epsilon)^{\frac{1}{N-1}} =1$. Further, with at least $1 -e^{-\Omega(N)}$ probability,
    $$
    \pi_{-i}^t(\mathbf 0) \le \left(\frac{1}{N-1}\sum_{j\ne i}\pi_j^0(0)\right)^{N-1} \le \frac{1}{N+1+2C+\epsilon}.
    $$ Combined with union bound, the statement can be deduced.
\end{proof}

\textbf{Proof for Theorem~\ref{theorem:good-initialization}}.
\textit{
    For $\mu\in\{\mathbf0, \mathbf1\}$, when the same preference preserved algorithm is applied, we must have $S_\text{SP} \subseteq S_\mu$. Furthermore, learning against fixed $\mu$ strictly enlarge the good initialization set as $S_\mu \backslash S_\text{SP} \ne \varnothing$.
}

\begin{proof}
    The proof for $\mu=\mathbf0$ is clear since any policy values $0$ more than $1$ and thus must converge to the global optimal policy $\mathbf0$. Hence, we only consider the case of $\mu=\mathbf1$.
    When $\mu=\mathbf1$, we can show that $\forall i$,
    \begin{equation}
        \label{proof:criteria-for-mu}
        \pi_{-i}(\mathbf0) \ge \frac{1}{N+C} \Rightarrow Q_i(0) \ge Q_i(1)
    \end{equation}
    by substituting $\mu(\mathbf0)=0,\mu(\mathbf1)=1$ in Equation~\ref{lemma:deltaQ}, which gives
    \begin{equation}
        \label{proof:computation-for-mu}
        Q_i(0) - Q_i(1) = (N+C)\pi_{-i}(\mathbf0) - 1.
    \end{equation}
    We first prove $S_\text{SP} \subseteq S_\mu$. For any two sequences $\{\tilde \pi^0, \tilde\pi^1,...\}$ updated by SP and $\{\pi^0, \pi^1,...\}$ updated by playing against $\mu$ where $\tilde\pi^0=\pi^0$, we use induction to show that $\forall t, i, \tilde\pi_i^t(0) \le \pi_i^t(0)$ and $Q_i^t(0)-Q_i^t(1)\ge \tilde Q_i^t(0)-\tilde Q_i^t(1)$. Suppose that hold for all $t'\le t$. Thus $Q_i^t(0)-Q_i^t(1)\ge \tilde Q_i^t(0)-\tilde Q_i^t(1)$. From the ratio monotone updating rule~\ref{def:monotone-update}, we directly get $\tilde\pi_i^{t+1}(0) \le \pi_i^{t+1}(0)$.
    
    Now we show that initialization $\pi_i^0(0)=(N+C)^{-\frac{1}{N-1}}$ belongs to $S_\mu$ (which is direct from the fact that $Q^t_i(0)\ge Q^t_i(1)$ if all $\pi_i^t(0) \ge (N+C)^{-\frac{1}{N-1}}$) but not $S_\text{SP}$, and thus $S_\text{SP}$ is strictly contained by $S_\mu$. To show $\pi_i^0(0)=(N+C)^{-\frac{1}{N-1}}$ does not converge to $\mathbf0$, it suffices to show that $\forall t, i, \pi_{-i}^t(0) < \frac{1}{N}$, which can be shown by induction. Clearly $t=0$ satisfies this. Now suppose this statement is true for all $t'\le t$, at step $t$, we have $\forall i$,
    \begin{align*}
        &Q_{i}^t(1) \le Q_{i}^t(0) \\
        \Rightarrow& \pi_{-i}(\mathbf 0)(1+2C+\epsilon + \|\pi^t\|_1) \ge 1 \\
        \Rightarrow& \pi_{-i}(\mathbf 0)\left(N + 1+2C+\epsilon - (N-1)\pi_{-i}(\mathbf0)^{\frac{1}{N-1}} \right) \ge 1 \\
        \Rightarrow& \pi_{-i}(\mathbf 0)(N+1+2C+\epsilon) \ge 1 + (N-1)\pi_{-i}(\mathbf0)^{\frac{N}{N-1}} \\
        \Rightarrow& \pi_{-i}(\mathbf 0)(N+1+2C+\epsilon) \ge 1 + (N-1)\pi_{-i}(0) \\
        \Rightarrow& \pi_{-i}(\mathbf 0) \ge \frac{1}{2+2C+\epsilon} \ge \frac{1}{N} ~~(N\gg C\gg \epsilon)
    \end{align*}
    Therefore, if $\forall t, i, \pi_{-i}^t(0) < \frac{1}{N}$, we have $\forall t,i$, $Q_{i}^t(0) < Q_{i}^t(1)$, and hence from ratio increase rule \ref{def:ratio-increase} $\pi_{-i}^{t+1}(0) < \pi_{-i}^t(0)< \frac{1}{N}$, which finishes our induction.
\end{proof}

\section{Experiment Details}\label{app:detail}
\subsection{Matrix Games}

\paragraph{Team RPS game.} We use a simple categorical policy for all algorithms and use an SGD optimizer with learning rate $0.1$ to run policy gradient.
Each algorithm is trained for 30k steps.
For SP and FSP, we simply train the single agent for 30k steps.
For PSRO with and without reset, we run 30 iterations and the BR policy in each iteration is trained for 1k steps.
For FXP, we run 15 iterations and the main policy and counter policy in each iteration are both trained for 1k steps.
The self-play probability $\eta$ is set to $0.2$ and decays exponentially to 0 with a factor of 0.97.
We set the initial policy to Rock for all algorithms for better visualization.
Changing the initial policy to other policies like a random policy will only change the starting point of the learning dynamics but will not change the final convergence results.

\paragraph{Seek-attack-defend (SAD) game.} We first elaborate the rewards in a formal manner. For each team $t_c$, suppose the seeking rewards is $\hat R_{t_c}$ and reward-level is $L_{t_c}$. Let $a_1,...,a_N$ be team $t_c$'s actions, we define $L_{t_c}$ as:

\begin{equation*}
    L_{t_c} = \left\{
    \begin{aligned}
        &0, && \exists i,j\text{ s.t. }a_i,a_j\in\{0,...,A\}\text{ and }|a_i-a_j|>1, \\
        &\min_{i, a_i\in \{0,...,A\}} a_i, && \text{otherwise}.
    \end{aligned}
    \right.
\end{equation*}

After that, the seeking reward is
\begin{equation*}
    \hat R_{t_c} = \sum_{i, L_{t_c} \le a_i \le L_{t_c}+1 } a_i.
\end{equation*}
Let $b_1,...,b_N$ be opponent team's actions, the final reward $R_{t_c}$ is
\begin{equation*}
    R_{t_c} = \left\{
    \begin{aligned}
        &0, && \forall i, a_i\ne \textit{defend} \text{ and } \exists i,j \text{ s.t. } b_i=b_j=\textit{attack} \\
        &\hat R_{t_c}, && \text{otherwise}.
    \end{aligned}
    \right.
\end{equation*}
Thus the utility is defined as $U_{t_c} = R_{t_c} - R_{t_{-c}}.$

To optimize the policy $\pi$, we directly compute the $Q$-function $Q^t_i(a_i)$ for each agent $i$ and action $a_i\in \{0,...,A,\textit{attack},\textit{defend}\}$ with policy $\pi^t$ against some opponent $\mu^t$. In SP, FoReL, Neural Replicator, MWU, and CFR, $\mu^t=\pi^t$. In FSP, $\mu^t=\eta \pi^t + (1-\eta)\frac{\sum_{i=1}^t \pi^t}{t}$. In PSRO, online DO, $\mu^t$ is the meta-policy $\sigma$. In FXP, $\mu^t=\eta \pi^t + (1-\eta) \sigma$. $\eta$ is fixed to $0.3$ for both FXP and FSP. We further define $V^t = U(\pi^t, \mu^t)$ here.

For SP, FSP, PSRO, online DO, and FXP, the policy is updated by a step towards stepwise best policy $\zeta_i^t = \argmax_{a_i} Q_i^t (a_i)$, i.e., $\pi^{t+1} = (1-lr)  \pi^t + lr \zeta_i^t$. We use $lr=0.1$ throughout these algorithms. For FoReL, we compute the accumulated Q value ${R_\text{FoReL}}_i^t(a_i) = \sum_{i=1}^t lr_t Q_i^t(a_i)$ and update $\pi^{t} = \text{softmax} ({R_\text{FoReL}}_i^t)$. Here $lr_t=20/\sqrt{t}$. For Neural Replicator, $\pi^{t+1} = \pi^{t} + \Delta t \pi^{t}Q^t$ with $\Delta t=0.8$. For MWU, $\pi^{t+1} \propto \pi^{t} \text{softmax}(kQ^{t})$ with $k=10$. For $CFR$, we aggregate the regret ${R_\text{CFR}}_i^t(a_i) = \sum_{i=1}^t Q_i^t(a_i) - V$ and update $\pi^{t} \propto\left( {R_\text{CFR}}_i^t \right)^+$ All the parameters are fine-tuned to make the policy $\pi$ converge quickly and stably, and each iteration of FXP and PSRO is stopped when the policy plateaus.

To evaluate, we compare them head-to-head with three opponents
\begin{align*}
\mu_\textit{seek}=&\{A,A,...,A\},\\ \mu_\textit{attack}=&\{\textit{attack}, \textit{attack}, A, A, ..., A\},\\ \mu_\textit{defend}=&\{\textit{defend}, A, A, ..., A\},
\end{align*}
since the global NE can be represented by 
$$\frac14 \left( \mu_\textit{seek} + \mu_\textit{attack} + 2\mu_\textit{defend} \right)$$
The exploitability of $\pi$ is defined as 
$$\text{exploitability}(\pi)=\sum_{\mu \in \{ \mu_\textit{seek}, \mu_\textit{attack}, \mu_\textit{defend}\} } \max(0, U(\pi, \mu)).$$ We remark that we report the exploitability of the average policy $\bar \pi^t= \frac{1}{t}\sum_{i=1}^t \pi^t$ for FSP and CFR, and for FXP and PSRO, the meta policy $\sigma$ is used for evaluation.

\begin{figure}[t]
     \begin{subfigure}[b]{0.45\textwidth}
         \centering
         \includegraphics[width=0.9\textwidth]{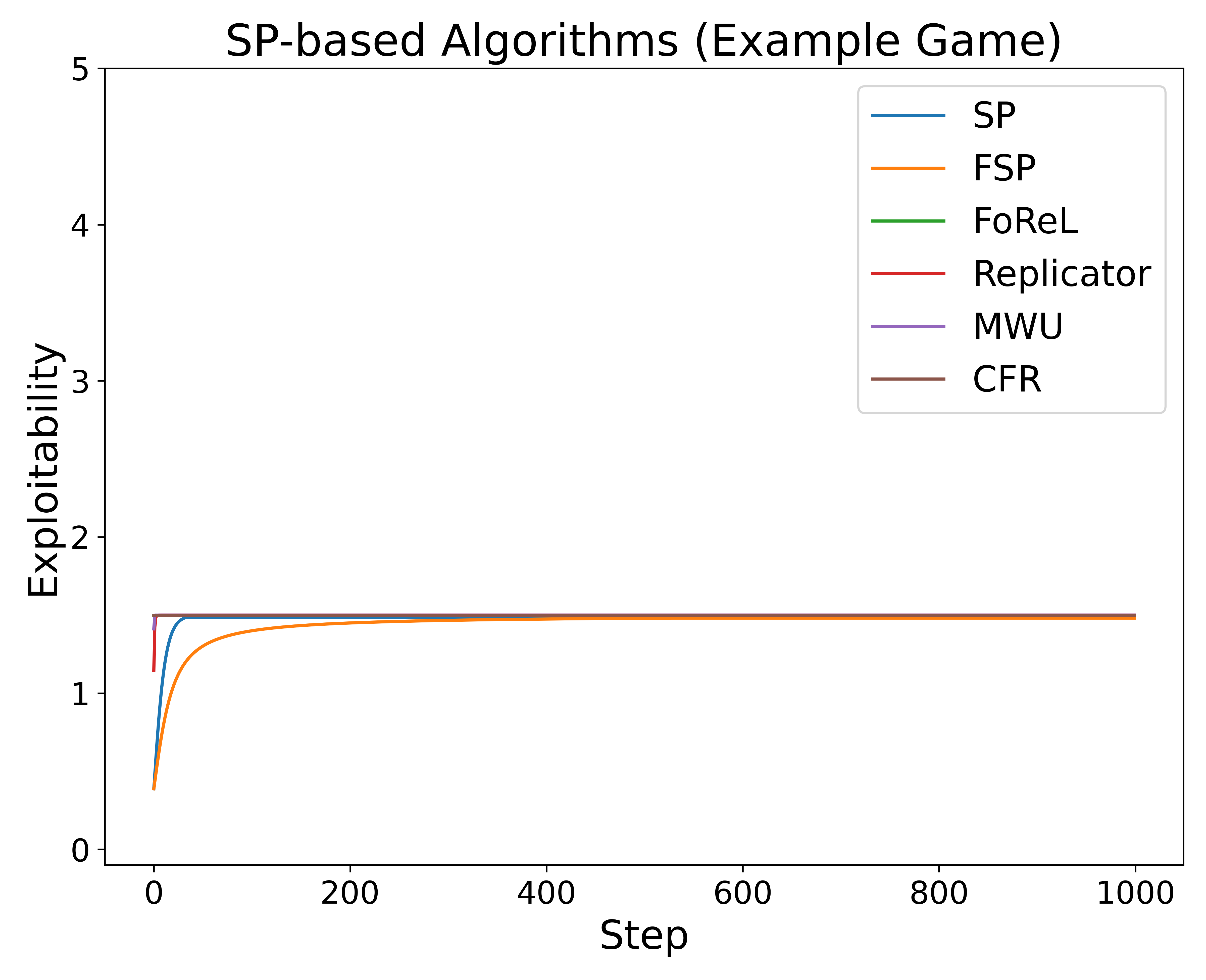}
         \caption{Exploitability of SP-based algorithms.}
         \label{fig:exp:exmaple-game-sp}
     \end{subfigure}
     \hfill
     \begin{subfigure}[b]{0.45\textwidth}
         \centering
         \includegraphics[width=0.9\textwidth]{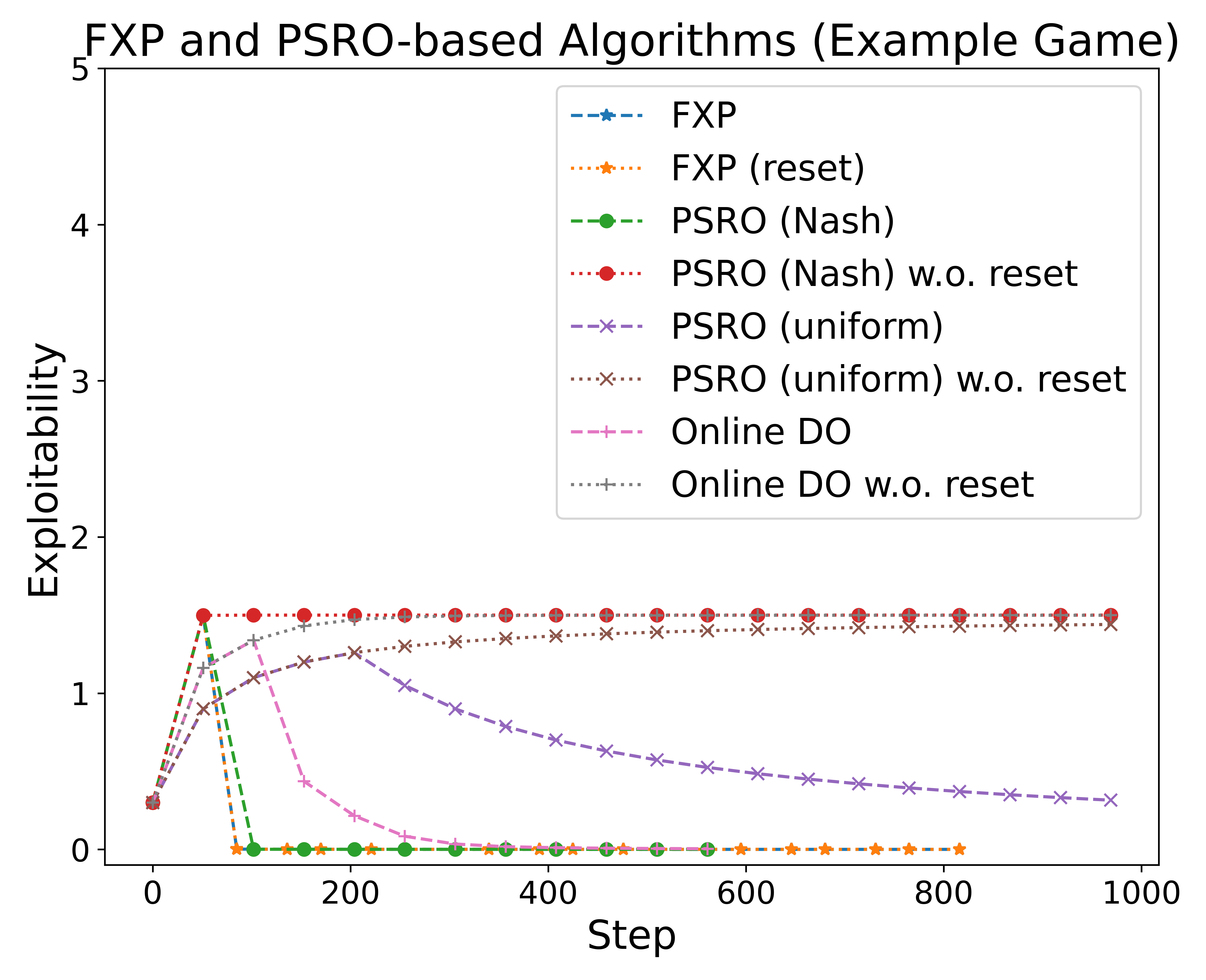}
         \caption{Exploitability of FXP and PSRO-based algorithms.}
         \label{fig:exp:example-game-psro}
     \end{subfigure}
    \caption{Results on the motivating example game.}
    \label{fig:exp:example-game}
\end{figure}

\paragraph{Motivating example game.}
We also run experiment on the motivating example game to validate our method. 
Following the descriptions in Section~\ref{sec:example}, we set $N=3, C=1.5, \epsilon=0.1$, and the self-play ratio $\eta$ is fixed to $0.3$. 
Each algorithm is trained for at most 1000 steps, and other algorithm setups are the same as the SAD game.
As shown in Figure~\ref{fig:exp:example-game}, FXP converge to the global NE after 85 steps, while PSRO$_\mathrm{Nash}$ uses 102 steps and Online DO uses 561 steps.
Other algorithms fails to converge to the global NE after 1000 steps, including PSRO$_\mathrm{Uniform}$ with 0.31 exploitability and SP-based algorithms with over 1.4 exploitabilities.

\subsection{MAgent Battle}

The environment of 3-vs-3 MAgent Battle is a gridworld with size $15\times 15$.
The observation of each agent is a state vector that includes the one-hot agent id, the position and hp of the agent itself, the position and hp of teammates, and that of the opponents.
The maximum length of an episode is set to 200.

All algorithms use a recurrent policy and are trained using 100M environment frames.
FSP saves a checkpoint every 1M frames and the self-play probability is 0.2.
PSRO and ODO are trained for 20 iterations and each iteration uses 5M frames to train the BR policy.
FXP is trained for 10 iterations and each iteration uses 5M frames for the main policy and 5M for the counter policy, the self-play probability is 0.2.
The approximate exploitability is estimated by training a BR of the learned policy or population for 20M environment frames using MAPPO. 
We use the standard Elo rating in evaluation, a difference of 100 points gives about 64\% win rate, and a difference of 200 points gives about 76\% win rate.
All training hyperparameters for different algorithms and BR learning are the same and listed in Table~\ref{tab:app:magent}.

\begin{table}[t]
\begin{tabular}{lc}
    \toprule
    Name & Value \\
    \midrule
    learning rate  & 5e-4 \\
    discount rate $\gamma$ & 0.99 \\
    GAE parameter $\lambda_{GAE}$ & 0.95 \\
    gradient clipping & 10 \\
    value loss coefficient & 0.5 \\
    entropy coefficient & 0.01 \\
    optimizer & Adam \\
    parallel threads & 100 \\
    chunk length & 10 \\
    PPO clipping & 0.2 \\
    PPO epoch & 5 \\
    MLP layer num & 3 \\
    MLP layer size & 64 \\
    LSTM layer size & 64 \\
    \bottomrule
    \\
\end{tabular}
\caption{Hyperparameters used in MAgent Battle environment.}
\label{tab:app:magent}
\end{table}

\subsection{Google Research Football}

We use the raw observation of GRF and construct a 292-dim vector as the observation input.
The vector contains information of the active player, ball, self team, opponent team, relative info, and game mode.
The detailed information is listed in Table~\ref{tab:app:fb_obs}. 
The action space of GRF contains 19 discrete actions including idle, move in 8 directions, pass, shot, sprint, slide, dribble, etc.

\begin{table}[t]
\begin{tabular}{cc}
    \toprule
    Length & Information \\
    \midrule
    21 & active player id, sticky actions \\
    5  & active player id, position, direction, tired factor \\
    3  & active player yellow card, red card, offside flag \\
    9  & ball position, direction, ownership \\
    55 & self team position, direction, tired factor \\
    33 & self team yellow card, red card, offside flag \\
    55 & opponent team position, direction, tired factor \\
    33 & opponent team yellow card, red card, offside flag \\
    3  & relative ball position, distance \\
    33 & relative self team position, distance \\
    33 & relative opponent team position, distance \\
    9  & game mode, goal difference, steps left \\
    \bottomrule
    \\
\end{tabular}
\caption{Information in the state vector of GRF.}
\label{tab:app:fb_obs}
\end{table}

We use a recurrent policy and run FXP with MAPPO on GRF full game with 100 iterations.
In each iteration, the main policy and counter policy are both trained for 20k model steps.
The self-play ratio $\eta$ is set to 0.2. 
All training hyperparameters for FXP in GRF are listed in Table~\ref{tab:app:fb}.

\begin{table}[H]
\begin{tabular}{lc}
    \toprule
    Name & Value \\
    \midrule
    learning rate  & 5e-4 \\
    discount rate $\gamma$ & 0.999 \\
    GAE parameter $\lambda_{GAE}$ & 0.95 \\
    gradient clipping & 10 \\
    value loss coefficient & 1 \\
    entropy coefficient & 0.01 \\
    optimizer & Adam \\
    parallel threads & 1000 \\
    batch size & 3600 \\
    chunk length & 10 \\
    PPO clipping & 0.2 \\
    PPO epoch & 10 \\
    MLP layer num & 4 \\
    MLP layer size & 256 \\
    LSTM layer size & 256 \\
    \bottomrule
    \\
\end{tabular}
\caption{Hyperparameters used in GRF environment.}
\label{tab:app:fb}
\end{table}

\subsection{Ablation Studies}

\begin{figure}
    \centering
    \includegraphics[width=0.4\textwidth]{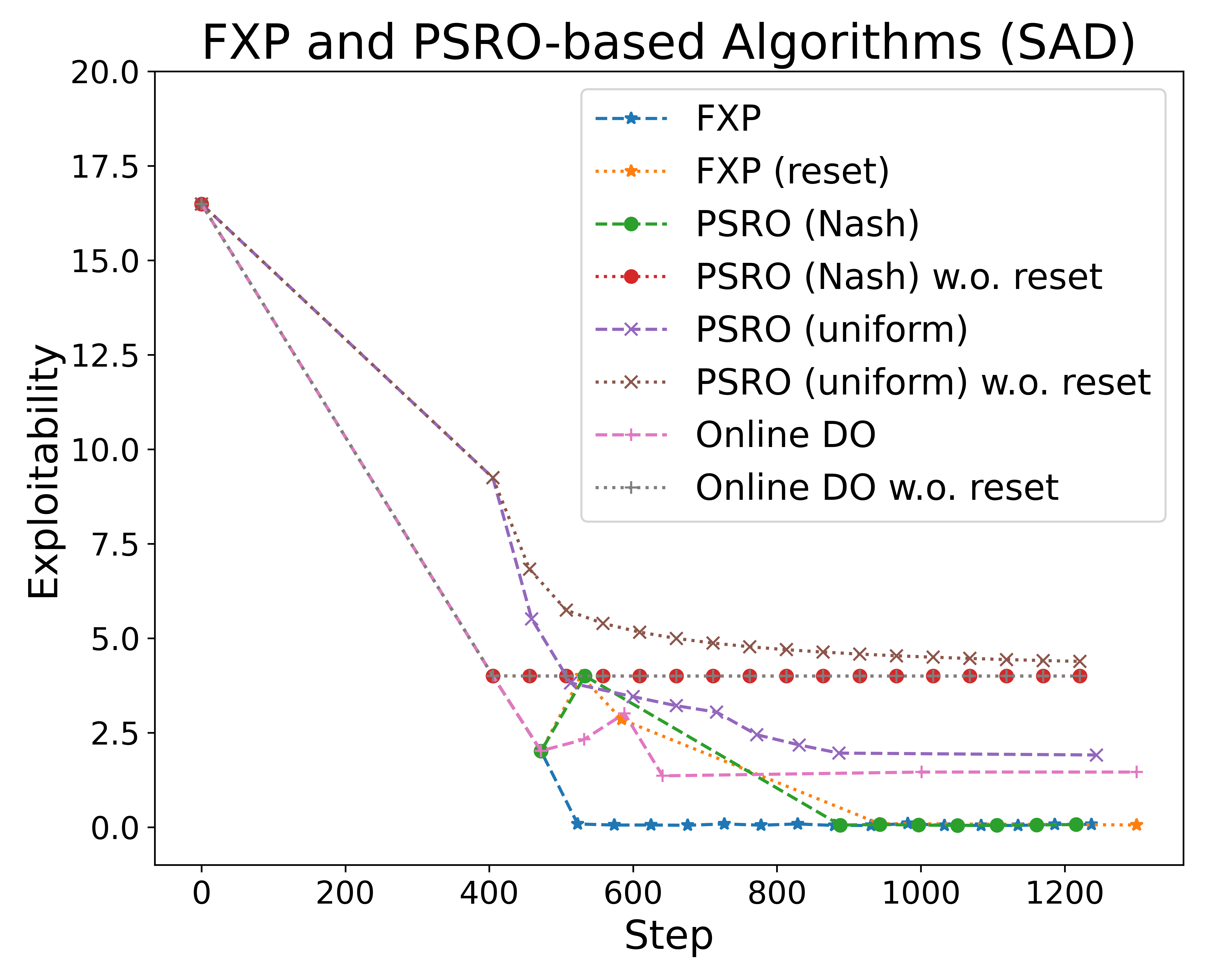}
    \caption{In SAD game, FXP w.o. reset converges slower than FXP w. reset.}
    \label{fig:exp:SAD-fxp-reset}
\end{figure}

\paragraph{FXP with and without reset.}
Intuitively, training from scratch may avoid biases in previous training, but will also greatly hurt the efficiency as shown in PSRO experiments. We run ablations for FXP and find the same results listed in Figure~\ref{fig:exp:SAD-fxp-reset}. In SAD game, FXP w.o. reset needs 523 steps to converge to a global NE while FXP w. reset needs 943 steps.

\begin{figure}
    \centering
    \includegraphics[width=0.45\textwidth]{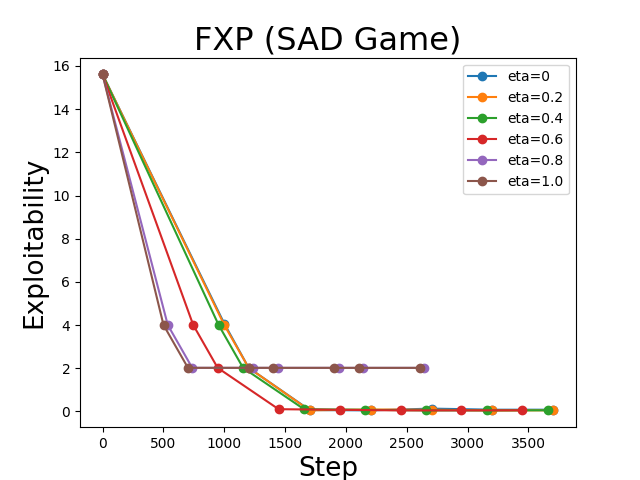}
    \caption{Results for FXP with different $\eta$.}
    \label{fig:exp:SAD-FXP-eta}
\end{figure}

\paragraph{Self-play ratio $\eta$.}
In general, larger eta leads to faster convergence but may converge to a local NE, and smaller eta is more likely to converge to a global NE with a slower speed. 
We conduct an ablation study in SAD games and find that eta=1 or 0.8 converge to a local NE very fast, while eta=0.6, 0.4, 0.2, 0 converge to the global NE and eta=0.6 uses the least steps. The results are shown in Figure~\ref{fig:exp:SAD-FXP-eta}.

\balance
\section{Comparison with AlphaStar}\label{app:compare}
Our work is different from AlphaStar~\cite{vinyals2019grandmaster} in the following ways.

1. AlphaStar tackles StarCraft II using a centralized policy that controls all units, which makes it a two-player zero-sum game without the local NE issue. We follow decentralized policies on mixed cooperative-competitive games, where local NE issue does exist since multiple agents in the same team choose actions in a decentralized fashion.

2. AlphaStar extends FSP with population-based training by maintaining a population of 12 different agents including main and different exploiters for all 3 races, while we only train a pair of main and counter policies. AlphaStar can be complementary to our work in the sense that we can scale up FXP further to train multiple pairs of policies and perform meta-optimization using population-based training.

3. There are also technical differences. On opponent sampling, we use a general meta-solver which can be, but is not limited to, win-rate-based prioritized sampling. It can also be uniform and Nash solvers as used in our matrix games experiments. AlphaStar starts from a behavior clone model from human data which is already rated as top 16\% players. It also uses statistics from human data to explicitly encourage diverse plays. We start from random models without human data and achieve strong results in challenging 11-vs-11 GRF environment.

4. AlphaStar is a practical work and gives intuition for using exploiters to benefit training, while we give a more in-depth analysis on why counter population can help policies get out of local NEs in mixed cooperative-competitive games. We also provide illustrative examples to show the effect of counter population, e.g., learning dynamics of the team RPS game in Figure~\ref{fig:exp:trps} and behavior analysis of MAgent Battle in Figure~\ref{fig:exp:magent}.


\end{document}